%% file: iclr2027_conference.tex
\documentclass{article}
\usepackage{iclr2027_conference,times}
\iclrfinalcopy
\input{math_commands.tex}

\usepackage{hyperref}
\usepackage{url}
\usepackage{booktabs}
\usepackage{graphicx}
\usepackage[table]{xcolor}
\usepackage{enumitem}
\usepackage{subcaption}
\usepackage{makecell}
\usepackage{pifont}

\title{Rvlm: Fewer Tokens, Better Resolution Scaling, Fine-Grained Spatial Grounding}

\author{Savya Khosla$^{1,2}$ \quad Chris Dongjoo Kim$^{1}$ \quad Alex Schwing$^{1}$ \\
  {\bf Derek Hoiem}$^{1}$ \quad {\bf Christopher Clark}$^{2}$ \quad {\bf Sangho Lee}$^{2}$ \\[0.5em]
  $^{1}$University of Illinois Urbana-Champaign \quad $^{2}$Allen Institute for AI
}

\newcommand{\rvlm}{RVLM}
\newcommand{\pvlm}{PVLM}
\newcommand{\boldheader}[1]{\noindent\textbf{#1.}}
\newcommand{\tval}[1]{\begin{tabular}[t]{@{}c@{}}#1\end{tabular}}
\newcommand{\diff}[2]{%
  \begin{tabular}[t]{@{}c@{}}
    #1\\[-1pt]
    \textcolor{green}{\scriptsize\bfseries (\(\downarrow\)#2\%)}
  \end{tabular}%
}
\newcommand{\perfgain}[2]{%
  \begin{tabular}[t]{@{}c@{}}
    #1\\[-1pt]
    \textcolor{green}{\scriptsize\bfseries (\(\uparrow\)#2)}
  \end{tabular}%
}

\newcommand{\perfsame}[1]{%
  \begin{tabular}[t]{@{}c@{}}
    #1\\[-1pt]
    \textcolor{black!60}{\scriptsize\bfseries (0.0)}
  \end{tabular}%
}
\newcommand{\cropgap}{\hspace{1.0em}}
\newcommand{\cmark}{\textcolor{tickgreen}{\ding{51}}}
\newcommand{\xmark}{\textcolor{crossred}{\ding{55}}}

\definecolor{green}{RGB}{0,130,90}
\definecolor{blue}{RGB}{200, 230, 242}
\definecolor{tickgreen}{RGB}{0,130,90}
\definecolor{crossred}{RGB}{190,55,55}

\definecolor{editred}{HTML}{D62728}
\definecolor{editblue}{HTML}{1768AC}
\definecolor{editpurple}{HTML}{9C27B0}
\newcommand{\rone}[1]{#1}
\newcommand{\rtwo}[1]{#1}
\newcommand{\ronetwo}[1]{#1}

\begin{document}

\maketitle
\lhead{Technical report}

\input{sections/0-abstract}
\input{sections/1-intro}
\input{sections/2-related}
\input{sections/3-method}

\input{sections/4-experiments}
\input{sections/5-conclusion}

\bibliography{iclr2027_conference}
\bibliographystyle{iclr2027_conference}

\input{sections/6-appendix}

\end{document}

%% file: math_commands.tex
\usepackage{amsmath,amsfonts,bm}

\def\eqref#1{equation~\ref{#1}}

\def\1{\bm{1}}

\DeclareMathAlphabet{\mathsfit}{\encodingdefault}{\sfdefault}{m}{sl}
\SetMathAlphabet{\mathsfit}{bold}{\encodingdefault}{\sfdefault}{bx}{n}



%% file: sections/0-abstract.tex
\begin{abstract}
Vision-Language Models (VLMs) tokenize images using fixed spatial grids, so visual token count grows with pixel area rather than scene content. We propose to group neighboring patches based on their visual appearance, allowing tokens to represent coherent image regions rather than fixed spatial windows. We introduce \rvlm{}, a VLM trained to use regions rather than patches as its visual units. \ronetwo{Analyzing region construction for general-purpose VLMs, we find that regions must cover the full image, stay spatially connected during end-to-end training, and preserve fine structure such as small text, while semantic grouping adds little.} \rvlm{} therefore partitions images using training-free Felzenszwalb-Huttenlocher segmentation, assigns each ViT patch to a region, and pools patches within each region into a variable-length sequence of visual tokens. The patch-to-region assignment is retained to support spatially precise prediction through coarse-to-fine decoding. Across tasks spanning VQA, pointing, counting, segmentation, and depth estimation, \rvlm{} matches or improves example-weighted task performance while using 21.5\% fewer visual tokens than an identically trained patch baseline. The advantage grows with image resolution, reaching a 5.3-point gain while using 51.8\% fewer visual tokens at the highest-resolution setting. The gains are largest on spatially demanding tasks, where prior token-reduction methods degrade most sharply. \rvlm{} adds no model parameters and improves inference throughput by up to 35.9\%. \rtwo{On unseen pathology and satellite images and on video benchmarks without any video training, \rvlm{} performs on par with the patch baseline while using 2--3$\times$ fewer visual tokens.} These results establish regions as a scalable alternative to patch-based visual representations while preserving the spatial precision required for grounded prediction.
\end{abstract}

%% file: sections/1-intro.tex
\section{Introduction}
\label{sec:introduction}

Tokenizing images into a uniform grid of patches is a standard design choice in Vision-Language Models (VLMs). However, patch tokenization ties token count to pixel count rather than scene content. As a result, a high-resolution image of a single object requires more tokens than a lower-resolution image of a cluttered scene, and the blank margins of a document may consume more tokens than the text, simply because they occupy more area. This content-agnostic allocation wastes the visual token budget and limits the maximum image resolution a patch-based VLM can process within a fixed token budget.

Existing models reduce visual token count through fixed spatial pooling~\citep{MiniGPTv2,InternVL2,Qwen2VL} or by pruning and merging patch tokens~\citep{ToMe,FastV,VisionZip}. Fixed pooling ignores image content, while pruning and merging can remove spatial coverage or obscure the correspondence between visual tokens and image locations needed for spatial grounding. Content-adaptive tokenization offers an alternative by letting visual units follow image content rather than a fixed grid~\citep{DiVT,EPOC,TREN}. However, existing approaches rely on learned features or segmentation models, introducing additional computational costs or making region construction sensitive to VLM training (Section~\ref{sec:constructing-regions}). 
\ronetwo{We find that region tokens depend less on semantic grouping than on whether regions cover the full image, stay spatially connected during end-to-end training, and preserve fine structure such as small text.}

We present \rvlm{}, a VLM trained to use regions rather than patches as its visual units. Unlike fixed spatial pooling, we group neighboring patches based on their visual appearance, forming more meaningful units for pooling visual information. We construct such regions using training-free Felzenszwalb-Huttenlocher segmentation~\citep{FH}, assign each ViT patch to a region, and pool patch features within regions to produce a visual token sequence. \rvlm{} maintains the patch-to-region assignment to support spatially grounded prediction. For pointing, it decodes locations coarse-to-fine, first predicting the relevant region, then a patch within that region, and finally a location within the patch. Tokenization adds no parameters over the patch baseline, and the coarse-to-fine decoding preserves spatial precision even when a single token represents a large portion of the image.

\begin{figure}[t]
    \centering
    \begin{subfigure}{0.38\linewidth}
        \centering
        \includegraphics[width=\linewidth]{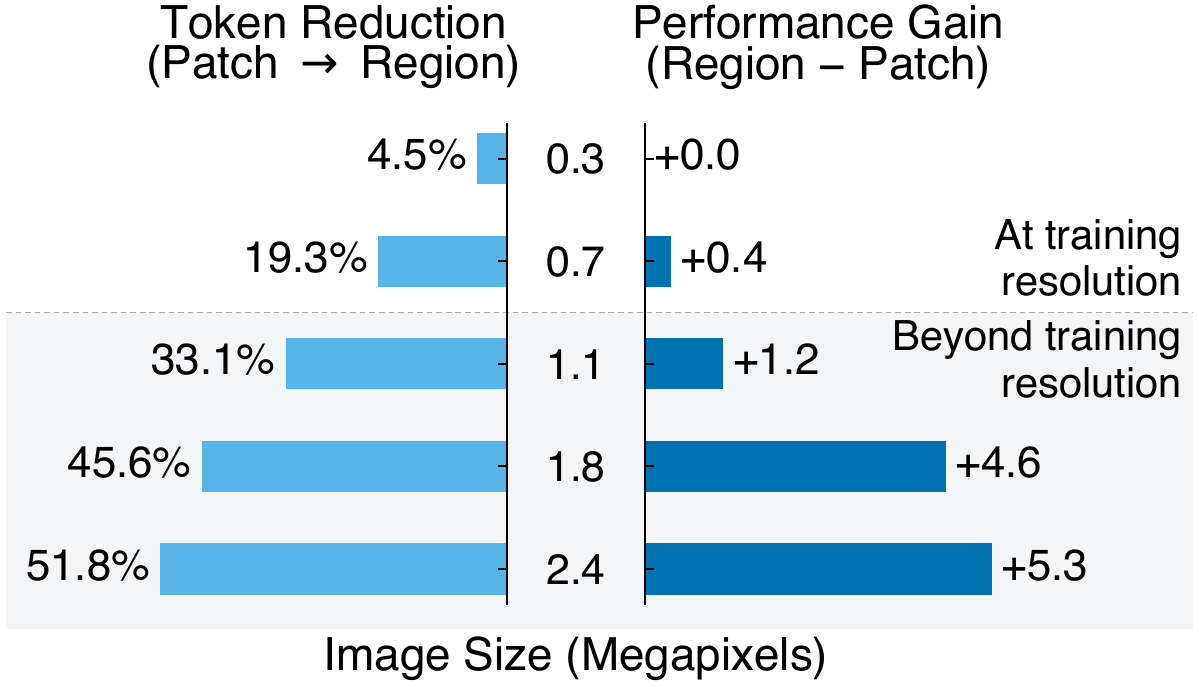}
        \caption{Regions scale better than patches as image size increases. \rvlm{} matches or outperforms an identically trained patch baseline while using fewer tokens. The largest gains emerge under test-time resolution scaling (Table~\ref{tab:native-res-bucketted-results}).}
        \label{fig:fixed-res}
    \end{subfigure}
    \hfill
    \begin{subfigure}{0.33\linewidth}
        \centering
        \includegraphics[width=\linewidth]{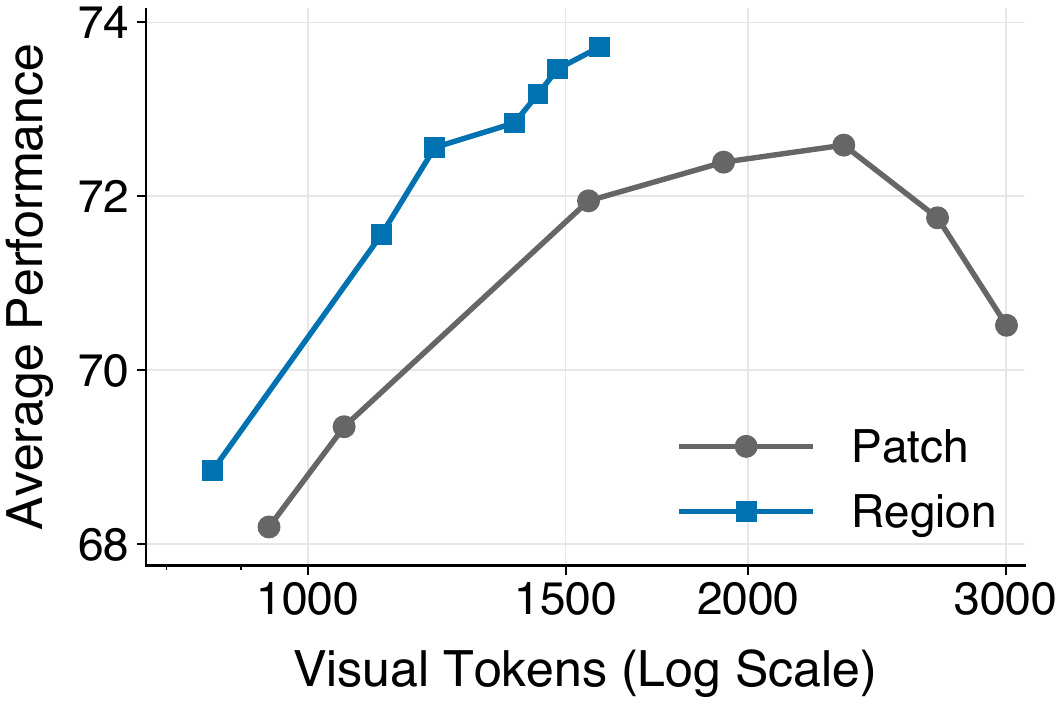}
        \caption{At a given token budget, \rvlm{} can process higher-resolution images than patch-based models, resulting in better performance. The gains are strongest on resolution-sensitive tasks.}
        \label{fig:fixed-tok}
    \end{subfigure}
    \hfill
    \begin{subfigure}{0.25\linewidth}
        \centering
        \includegraphics[width=0.93\linewidth]{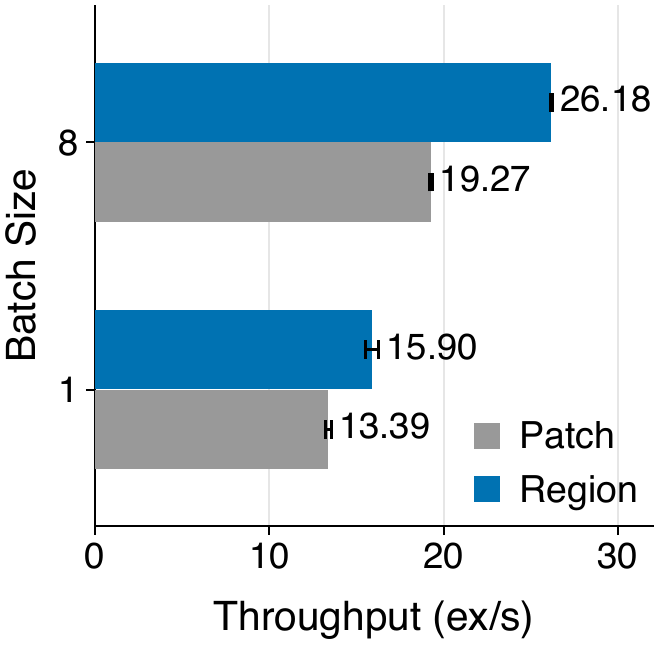}
        \caption{\rvlm{}'s shorter sequences increase inference throughput, with gains growing from 19\% at batch size 1 to 36\% at batch size 8 (Table~\ref{tab:inference-cost}).}
        \label{fig:inference-cost}
    \end{subfigure}

    \caption{{\bf Region tokens scale better with image resolution while improving inference efficiency.} See Section~\ref{sec:regions-vs-patches} for details.}
    \label{fig:teaser}
\end{figure}

We evaluate \rvlm{} across 35 tasks and compare it with a matched patch-based VLM and similarly sized open-weight models. Our results show that regions improve resolution scaling and inference efficiency while preserving the spatial correspondence needed for grounded prediction (Figure~\ref{fig:teaser}). We also study the properties required for effective region construction (Section~\ref{sec:constructing-regions}).

In summary, our main contributions are:
\begin{enumerate}[left=0pt]
    \item \ronetwo{{\bf What makes region tokens work for general-purpose VLMs.} Analyzing region construction strategies, we find that regions must cover the full image, stay spatially connected during end-to-end training, and preserve fine structure such as small text, while semantic grouping adds little. Segmenters that leave pixels uncovered cannot support pointing at those locations, and regions derived from the VLM's own features fragment as the encoder trains. In contrast, learned segmenters and clustering on a separate frozen encoder do not outperform classical Felzenszwalb-Huttenlocher segmentation, which is fast, adds no model parameters, and preserves small text better than SLIC.}
    \item {\bf An efficient VLM built on region tokens.} We present \rvlm{}, a VLM that replaces fixed spatial pooling with content-adaptive region-based tokenization. \rone{Retaining the patch-to-region correspondence lets \rvlm{} reduce visual tokens while matching or improving grounding.} Compared to an identically trained patch-based VLM, \rvlm{} uses 21.5\% fewer visual tokens on average and improves inference throughput by up to 35.9\%. It also scales more effectively to higher test-time resolutions, achieving a 5.3-point gain with 51.8\% fewer tokens at the highest resolution tested. \rone{Its gains are largest on spatial tasks such as counting, segmentation, and depth, where prior methods that reduce token count degrade sharply.} \rtwo{Making region tokens work for dense outputs requires adapting the pooling, position encoding, and pointing and segmentation decoders to irregular regions. Naively replacing patch tokens with region tokens collapses coarse-to-fine pointing on PixMo-Point to 10.0, compared with 84.4 for patch tokens and 86.6 with our design (Table~\ref{tab:dense-outputs}).}
\end{enumerate}

%% file: sections/2-related.tex
\section{Related Work}
\label{sec:related}

\boldheader{Visual tokenization in VLMs}
VLMs encode images into patch features using a ViT and map them into the LLM embedding space through a connector~\citep{LLaVA,PaLI}. Since patch sequences grow rapidly with image resolution, connectors typically reduce the number of visual tokens passed to the LLM. One approach uses a fixed number of learned queries~\citep{BLIP2,Flamingo,QwenVL}, decoupling token count from image size but weakening the correspondence between tokens and image locations. A more common approach pools patch features on a fixed spatial lattice. Existing methods use pixel shuffle~\citep{InternVL,Idefics3,SmolVLM}, adaptive average pooling~\citep{DeCo}, convolutional or deformable projectors~\citep{Honeybee,MobileVLMv2,DocOwl}, and cross-attention over local patch features~\citep{TokenPacker,Molmo,Molmo2}. While these designs differ in how they aggregate features, their spatial pooling units are determined by position rather than image content. Increasing the pooling window reduces the token count but also coarsens the visual representation, potentially impairing localization~\citep{SmolVLM}. In contrast, \rvlm{} constructs pooling units from image content, producing region tokens of varying spatial extent while retaining explicit patch-to-token correspondence for spatial grounding.

\boldheader{Content-adaptive tokenization}
Another line of work constructs content-adaptive visual tokens via semantic clustering of the vision encoder's patch features~\citep{ChatUniVi,SeTok,DiVT,Granulon,REN,TREN}. These approaches, however, depend on the spatial fidelity of the patch features, and we find that end-to-end VLM training destabilizes region assignments, producing spatially disconnected groups (Section~\ref{sec:constructing-regions}). Related adaptive-length tokenizers represent images as variable-length latent sequences~\citep{KARL,ElasticTok}, but lack explicit correspondence between tokens and image locations, limiting their use for spatial grounding. In contrast, \rvlm{} constructs spatially connected regions directly from image pixels while retaining patch-to-region correspondence for precise spatial grounding. Prior work has also explored region-based visual tokenization in vision encoders, image autoencoders, and VLMs~\citep{SuiT,Reader,Groma,SAMProjector}. \rone{Like most region-based representations~\citep{RegionRevisited,REN,TREN,EPOC,DiVT}, \rvlm{} forms region tokens from the features of a pretrained patch encoder, reusing strong encoders while changing the unit the LLM reasons over.} Closest to ours, \citet{EPOC} study SAM-based region tokenization and propose a lightweight learned boundary detector followed by watershed segmentation to ensure complete image coverage. However, learned detectors and segmenters introduce additional computational and memory overhead during region construction. \rone{\rvlm{} instead uses training-free segmentation that depends only on the input image, so regions are available before the forward pass without additional model parameters.
In our implementation, segmentation runs on the CPU during data loading, an engineering choice that keeps it off the accelerator and lets it overlap with model execution.}

\boldheader{Post-hoc token compression}
A separate line of work reduces visual tokens by pruning less important patches~\citep{VisPruner,PruMerge,FastV,SparseVLM,PyramidDrop,VisionZip} or merging redundant features~\citep{ToMe,DART,DivPrune,CDPruner}. Although effective for image-level understanding, these methods degrade performance on spatially grounded tasks~\citep{RightProblem,GAP}. Pruning removes information at discarded locations, while merging can obscure the correspondence between tokens and individual patches. In contrast, \rvlm{} constructs content-adaptive pooling units rather than compressing an existing token sequence, retaining complete patch coverage and spatial correspondence. We compare these approaches at matched per-image token budgets in Section~\ref{sec:token-compression}.

%% file: sections/3-method.tex
\section{Method}
\label{sec:method}

\begin{figure}[t]
    \centering
    \includegraphics[width=\linewidth]{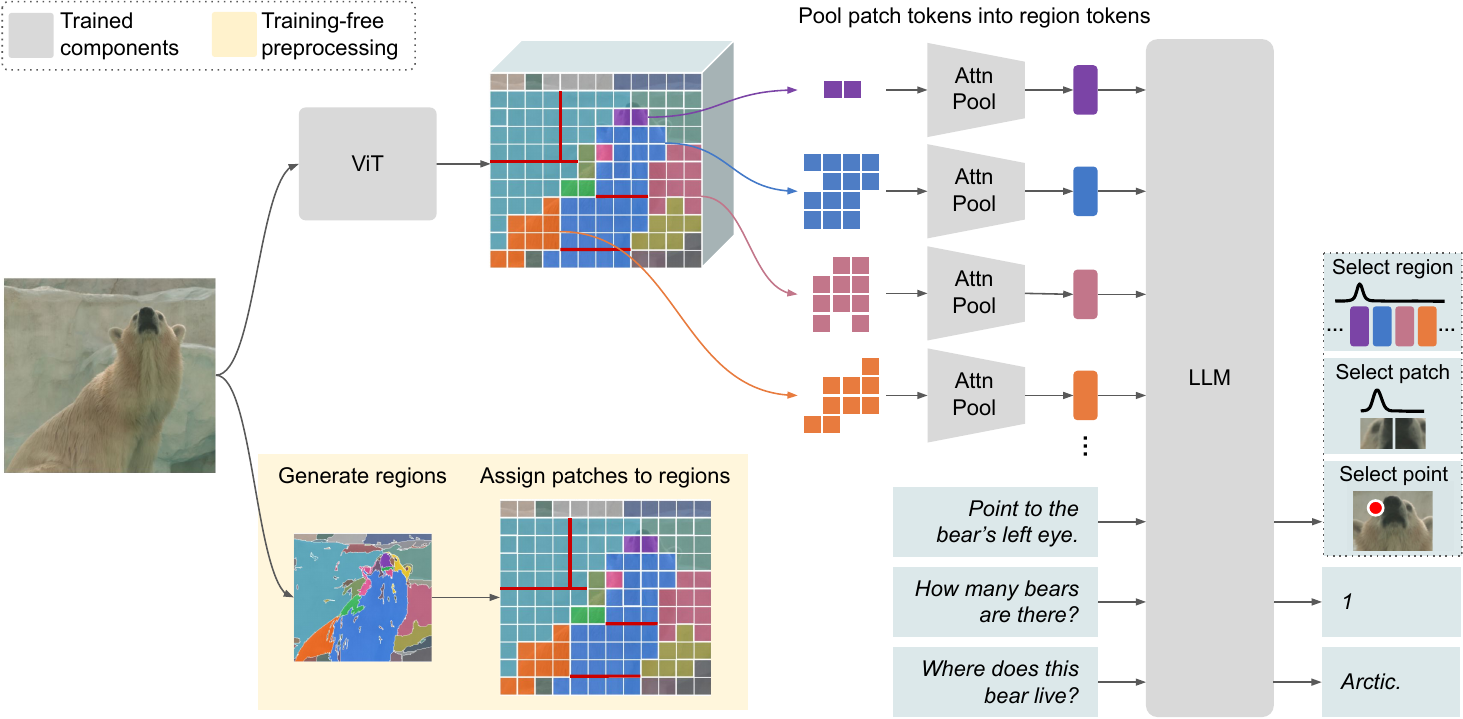}
    \vspace{-1.6em}
    \caption{{\bf Overview of \rvlm{}.} We generate content-adaptive regions using Felzenszwalb-Huttenlocher segmentation and assign each ViT patch to a region. This is a CPU-side preprocessing that runs in parallel to the model execution. Patch features within each region are aggregated through attention pooling to produce a sequence of region tokens for the LLM. The retained patch-to-region correspondence enables coarse-to-fine spatial grounding, where the LLM selects a region, then a patch within it, and finally a precise point within the patch.}
    \label{fig:method-overview}
\end{figure}

We introduce \rvlm{}, a VLM trained to use regions as its visual units (Figure~\ref{fig:method-overview}). Conventional spatial pooling summarizes patches within fixed spatial windows, regardless of their visual content. Instead, we group spatially connected patches with similar appearance and summarize each group into a visual token. These groups need not correspond to complete objects \ronetwo{(as argued for sub-object tokens by \citet{EPOC})}; when an object spans multiple regions, the LLM can combine information across their tokens, just as it does across patch tokens or subword tokens. Our goal is therefore to construct content-adaptive pooling units that follow local image structure while covering every pixel of the image.

\subsection{Modeling}
\label{sec:modeling}
\rvlm{} follows the standard VLM architecture, combining a pretrained vision encoder and LLM through a trainable connector. We initialize the vision encoder from SigLIP2-So400m/14~\citep{SigLIP2} and the LLM from Qwen3-4B-Instruct~\citep{Qwen3}. The connector is initialized from scratch, and it aggregates ViT patch features into region tokens using multi-head cross-attention. Following Molmo~\citep{Molmo}, the vision encoder processes each image as a low-resolution overview and up to $K$ overlapping high-resolution crops. If the image requires more than $K$ crops, we downscale it until it fits within this budget. The low-resolution overview is processed through standard $2{\times}2$ patch pooling, producing 196 visual tokens. For the high-resolution pathway, we extract ViT features from individual crops and assemble them into a common patch lattice before applying region-based attention pooling. All spatially grounded predictions, including pointing and segmentation, operate on this high-resolution representation, while the overview is used only for global context. We use $K{=}14$ during training and increase it to $K{=}32$ at inference to study scaling to higher image resolutions.

\boldheader{Generate regions}
We use Felzenszwalb-Huttenlocher (FH) segmentation~\citep{FH}
to partition images into spatially connected regions with complete
coverage. FH merges neighboring components based on color similarity
relative to their internal variation. We use a scale parameter $k{=}100$, which controls how aggressively merges are allowed; Gaussian smoothing with $\sigma{=}0.8$ before edge weighting, which reduces noise; and a minimum component size of 32 pixels, below which components are absorbed into neighbors. The image's longer side is capped at $512$ pixels during training and $768$ during inference; the resulting label map is resized using nearest-neighbor interpolation.
\rone{We run FH on the CPU in the dataloader workers, so that it does not compete with the model for GPU time.} 
We discuss other region generation strategies in Section~\ref{sec:constructing-regions}.

\boldheader{Assign patches to regions}
We assign each ViT patch to exactly one FH region. To preserve fine structures, we assign each patch to the smallest region covering at least 25\% of its pixels, falling back to the region with the largest overlap if none meets this threshold. For efficient batched attention pooling, we limit each group to $w{=}16$ patches, splitting larger regions into spatially contiguous groups. This retains all patch features for spatial grounding while controlling padding and the spatial extent of each token. We study the choice of $w$ in Appendix~\ref{app:connector-width-ablation}.
\rtwo{Both rules are required for grounding: splitting keeps every patch selectable in the second stage of coarse-to-fine pointing, which considers a fixed number of patches per region token, and exhaustive, exclusive assignment makes every target reachable and unambiguous.
Without them, coarse-to-fine pointing collapses (Appendix~\ref{app:dense-outputs}).}

\boldheader{Pool patches into region tokens}
We convert each patch group into a visual token using a cross-attention connector. For each group, we gather features from the third-to-last and ninth-from-last layers of the vision encoder and concatenate them. To capture the spatial structure of each region, we add absolute sinusoidal position embeddings encoding the spatial coordinates $(t,h,w)$ of each patch. Following \citet{Molmo2}, we use the mean of the valid patch features as the query for a multi-head cross-attention layer, with the individual patch features serving as keys and values. Groups smaller than $w{=}16$ are padded, and the padded positions are masked during attention. The resulting representation is projected into the LLM embedding space through an MLP, producing one visual token per group. Unlike fixed-window pooling, this connector adaptively aggregates features from spatially contiguous regions of varying sizes.
\rone{Notably, in our setup, pooling does not discard patch-level features: the grounding heads can read the unpooled ViT patch features, which we retain with the patch-to-region assignment (Appendix~\ref{app:task-formulation}).}

\boldheader{Position encoding}
We position each region token in the LLM using interleaved M-RoPE~\citep{MRoPE}. Specifically, we assign each token a 3D position $(t,h,w)$, where $t$ denotes the image's position in the input sequence and $(h,w)$ are the coordinates of the patch closest to the region's spatial center. This places region tokens in the same spatial coordinate system as the original patch grid without requiring regions to be rectangular. Thus, M-RoPE encodes the region's position within the image for the LLM, while the spatial structure of the region is encoded using the absolute patch-level positional embeddings added before the attention pooling of patch groups into region tokens.

\boldheader{Coarse-to-fine spatial grounding}
For pointing, we follow the coarse-to-fine decoding strategy of MolmoPoint~\citep{MolmoPoint}, adapted to region tokens. Instead of directly predicting image coordinates, the LLM emits a special token and uses its hidden state as a query that is scored against keys for all region tokens (dot-product retrieval); the highest-scoring region is selected. A second special token likewise retrieves a ViT patch among that region’s owned patches, and a third predicts the point’s location within the selected patch on a $4{\times}4$ grid. We use the patch-to-region assignment to map the predicted region, patch, and local position back to image coordinates. \ronetwo{This decoding strategy lets our region-based VLM ground precisely, which would otherwise be challenging because a region can span a large part of the image (Appendix~\ref{app:dense-outputs}).} For segmentation, we reuse the first stage of pointing to select an anchor region for the target object. The LLM then generates a special segmentation token whose hidden state is used to predict mask scores at three levels: regions, patches within each region, and $4{\times}4$ cells within each patch. For each cell, we sum its region-level, patch-level, and cell-level scores to obtain a mask logit. Using the retained patch-to-region assignment, we map these logits back to their corresponding image locations, apply a sigmoid, and threshold them at 0.5 to obtain a binary mask. This allows irregular region tokens to support dense, spatially precise predictions without requiring a regular pooling grid.

\subsection{Training}
We jointly train the vision encoder, connector, and LLM in two stages, following the Molmo2 training recipe~\citep{Molmo2}. Stage 1 consists of 31,971 steps of multimodal pretraining on captioning, pointing, counting, OCR, and text instruction data. Stage 2 consists of 24,000 steps of supervised fine-tuning, adding image QA, segmentation, and depth prediction. We optimize next-token cross-entropy together with the spatial grounding objectives of MolmoPoint~\citep{MolmoPoint}. Pointing uses cross-entropy at each level of coarse-to-fine decoding, while segmentation uses focal and Dice losses. Appendix~\ref{app:implementation-details} provides the full training configuration.

%% file: sections/4-experiments.tex
\section{Experiments}
\label{sec:experiments}

We evaluate whether region tokens can replace patches as the visual representation in general-purpose VLMs. To isolate the effect of visual tokenization, we train \pvlm{}, a Patch-based Vision-Language Model with the same architecture, training data, and recipe as \rvlm{}. The two models differ only in their visual tokenization strategy. We first compare both models against similarly sized open-weight VLMs (Section~\ref{sec:task-comparison}). We then evaluate the effect of region tokenization across 35 tasks spanning image QA, pointing, counting, segmentation, and depth estimation (Section~\ref{sec:regions-vs-patches}). Next, we compare \rvlm{} against post-hoc token compression methods (Section~\ref{sec:token-compression}). Finally, we analyze region generation strategies and identify the properties needed for effective region-based VLMs (Section~\ref{sec:constructing-regions}). \rtwo{Appendix~\ref{app:additional-ablations} adds design ablations and results on unseen domains, position encoding, model scale, text-only tasks, and video tasks.}

\subsection{Comparison with open-weight VLMs}
\label{sec:task-comparison}

\input{tables/task-comparison}

Table~\ref{tab:task-comparison} compares \rvlm{} and \pvlm{} with similarly sized open-weight VLMs across standard image benchmarks. Both models achieve competitive performance across image QA, pointing, and counting. Importantly, \rvlm{} achieves this performance using content-adaptive region tokens rather than conventional patch tokens. These results also establish that \pvlm{} is a strong baseline. Since \pvlm{} and \rvlm{} share the same architecture, training data,
and recipe, our subsequent comparisons isolate the effect of visual tokenization.

\subsection{Region vs.\ Patch Tokenization}
\label{sec:regions-vs-patches}
We compare \pvlm{} and \rvlm{} across 35 tasks spanning VQA, pointing, counting, segmentation, and depth (Table~\ref{tab:native-res-overall-results}). \rvlm{} improves the family-balanced average by 0.7 points and the example-weighted average by 0.6 points while using 21.5\% fewer visual tokens. This shows that regions shorten the visual sequence without sacrificing performance.

\boldheader{Regions improve performance on spatial tasks}
Table~\ref{tab:native-res-overall-results} shows that the largest gains are on counting ($+0.9$), segmentation ($+1.3$), and depth ($+1.9$), with particularly strong improvements on SA-Co Gold, DDAD, Waymo, and nuScenes. These gains indicate that regions are more effective representation units for reasoning about coherent objects and surfaces. The largest drops occur on InfoVQA ($-4.9$), ScreenSpot Pro ($-3.3$), and ChartQA ($-0.8$), where relevant information often occupies very small areas (e.g., small text in thin fonts). Failure analysis shows that these super-fine details are sometimes absorbed into larger surrounding regions and reduce spatial precision.

\input{tables/native-res-overall-results}

\boldheader{Regions scale better with image size}
We group examples by native resolution into five buckets to test how region and patch tokenization scale with image size. Figure~\ref{fig:fixed-res} summarizes the results, with full breakdown reported in Table~\ref{tab:native-res-bucketted-results}. As resolution increases, \rvlm{} achieves progressively larger performance gains while using increasingly fewer visual tokens than the patch baseline. Region tokenization therefore benefits more from additional image detail without the rapid sequence-length growth of patch tokenization. This makes it particularly well suited to high-resolution settings such as documents, screens, robotics, and dense visual prediction.

\boldheader{Regions yield better performance at fixed token budget}
We select high-resolution images from all tasks in the evaluation suite, yielding 21,273 examples ranging from 2.08-2.63 MP, and re-evaluate both \rvlm{} and \pvlm{} across seven ``maximum-resolution" settings. In each setting, the selected high-resolution examples are capped at a prescribed maximum image resolution, allowing us to measure performance as progressively more image detail is made available. This produces the matched token-budget comparison shown in Figure~\ref{fig:fixed-tok}. Since regions can represent higher-resolution inputs within the same visual token budget, \rvlm{} consistently outperforms \pvlm{} when the LLM receives a comparable number of visual tokens. This allows \rvlm{} to trade its token savings for higher input resolution and improve performance under a fixed LLM context budget.

\rtwo{\textbf{Regions generalize across domains, scales, and modalities.}
On unseen pathology and satellite images, \rvlm{} performs on par with \pvlm{} while using about half as many visual tokens (App.~\ref{app:unseen-domains}). On video, which neither model was trained on, \rvlm{} stays within 0.5 points of \pvlm{} with 3$\times$ fewer tokens (App.~\ref{app:text-video}). The same trends hold at the 1.7B scale (App.~\ref{app:model-scale}), and region tokenization leaves text-only performance essentially unchanged (App.~\ref{app:text-video}).}

\boldheader{Cost analysis}
We report end-to-end throughput and memory measurements in Table~\ref{tab:inference-cost}. The throughput gains with region tokenization come from shorter visual sequences processed by the LLM. FH segmentation adds roughly 49 ms of CPU preprocessing per image. With eight dataloader workers, this preprocessing keeps pace with GPU execution at both batch sizes\rone{, so the throughput in Table~\ref{tab:inference-cost} includes region construction without using accelerator time}. \rtwo{Further, FH's cost stays bounded on very high-resolution inputs because FH runs on a downscaled copy whose longer side is capped at 768 pixels.} A finer-grained analysis across image resolutions (Table~\ref{tab:inference-cost-sweep}) shows that \pvlm{} is slightly faster at one and four crops for batch size 1, where the savings from fewer tokens do not yet offset region-pooling overhead. From eight crops onward, \rvlm{} achieves lower latency at both batch sizes, with increasing latency and memory savings at higher resolutions. These results are measured with a 4B-parameter model; larger LLMs may yield greater efficiency gains from shorter visual sequences.

\input{tables/inference-cost}

\begin{figure}[t]
    \centering
    \includegraphics[width=\linewidth]{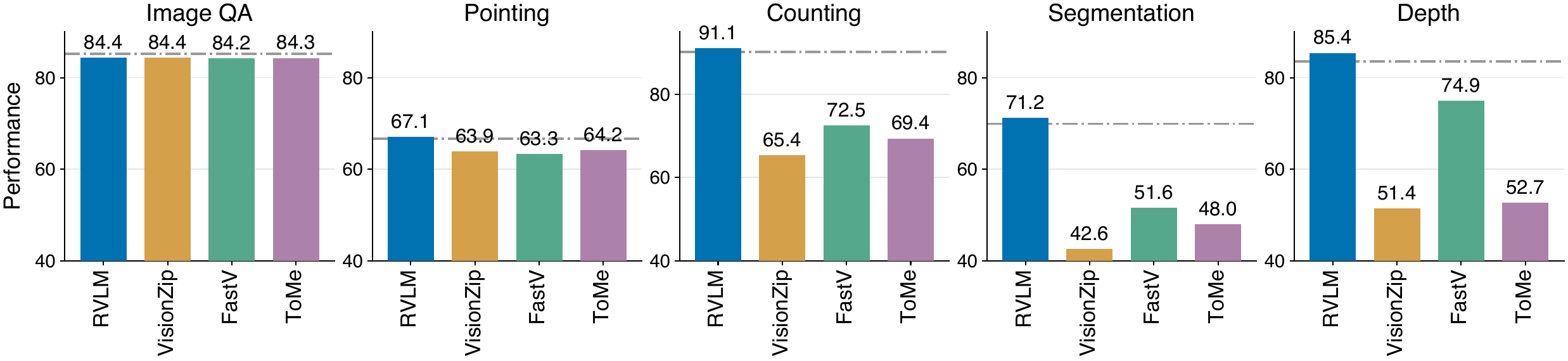}
    \caption{{\bf Post-hoc token compression methods degrade sharply on tasks requiring spatial precision.} \rvlm{} outperforms token pruning (FastV), token merging (ToMe), and hybrid (VisionZip) methods on pointing, segmentation, counting, and depth estimation, while all methods perform comparably on Image QA. Horizontal line denotes the \pvlm{} baseline.}
    \label{fig:compression-performance}
\end{figure}

\subsection{\rvlm{} vs.\ Token Compression}
\label{sec:token-compression}

Another way to shorten visual sequences is to compress patch tokens via pruning (FastV~\citep{FastV}), merging (ToMe~\citep{ToMe}), or both (VisionZip~\citep{VisionZip}). We apply these methods to \pvlm{} and compare them with \rvlm{} in Figure~\ref{fig:compression-performance}. For a controlled comparison, we match the visual token budget per image by setting each compression method's target token count to the number of region tokens produced by \rvlm{} for that image. This gives the baselines a per-image signal that their own formulations do not assume, enabling them to compress dense scenes less aggressively than sparse ones.

As shown in Figure~\ref{fig:compression-performance}, post-hoc compression largely preserves performance on image understanding tasks, consistent with prior findings that visual sequences contain redundancy. However, we observe large performance drops on pointing, segmentation, and point-to-depth prediction, which is consistent with the findings of \citet{GAP}. Pruning removes individual spatial representations, while merging combines information from multiple locations into shared tokens. In contrast, \rvlm{} retains an explicit assignment from every image patch to a region token, allowing its coarse-to-fine decoder to recover spatially precise predictions. The advantage of \rvlm{} is therefore not just fewer visual tokens, but a shorter sequence that preserves both spatial coverage and correspondence.

\subsection{Constructing Regions}
\label{sec:constructing-regions}

\begin{figure}[t]
    \centering
    \includegraphics[width=\linewidth]{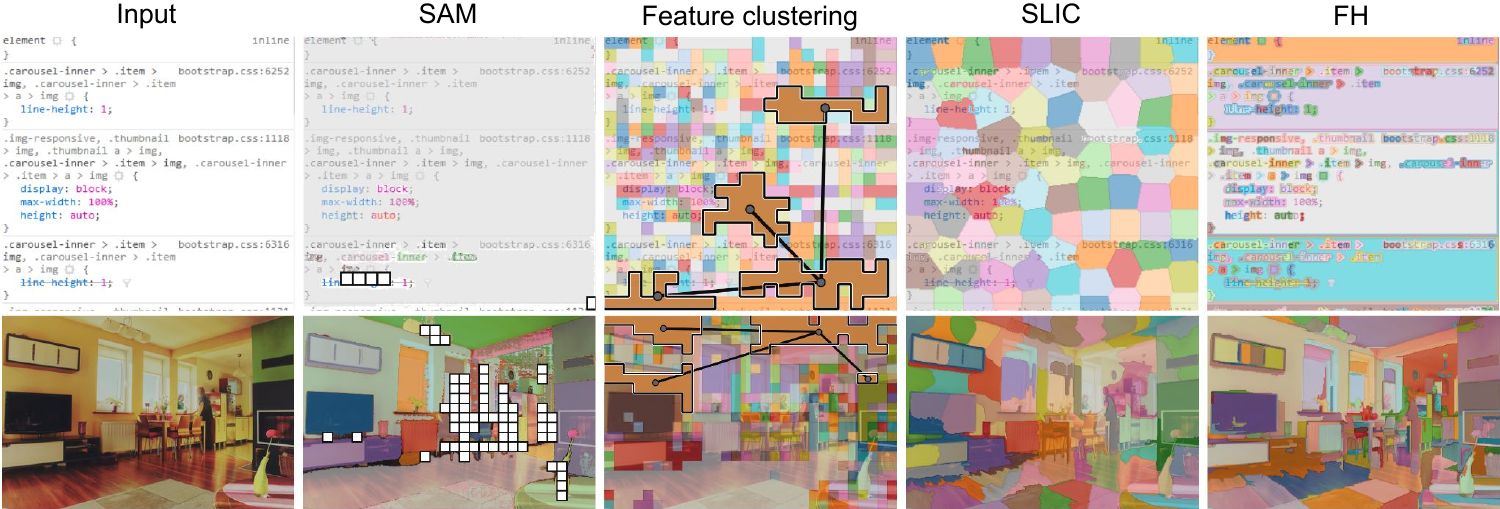}
    \caption{{\bf Comparing region generation methods.} SAM leaves image regions uncovered (white boxes), while learned feature-based methods (DiVT, T-REN) produce disconnected regions during VLM training. SLIC often fails to isolate text, whereas FH preserves fine text structures with connected regions and complete image coverage.}
    \label{fig:region-generation-methods}
\end{figure}

\rvlm{} requires mapping image patches to regions. We evaluate multiple region generation strategies and analyze their failure cases. \ronetwo{Our analysis shows that, for general-purpose VLMs, regions must cover the full image, stay spatially connected during end-to-end training, and preserve fine structure such as small text, while semantic grouping adds little.} We discuss each strategy below and summarize our findings in Table~\ref{tab:region-generation-methods} and Figure~\ref{fig:region-generation-methods}.

\boldheader{Region encoder network}
Our first approach replaces the patch-based ViT with T-REN~\citep{TREN}, an efficient image encoder that directly produces region tokens. While promising as a natural starting point for region-based VLMs, this approach reveals two fundamental challenges. First, the region quality depends strongly on the choice of vision encoder. T-REN relies on the spatial structure of dense patch features to generate regions, making some vision backbones better suited than others. As shown in Table~\ref{tab:region-generation-methods}, we train \rvlm{} using T-REN with SigLIP-2, DINOv3, and C-RADIOv4-H, and find that SigLIP-2-based T-REN lacks sufficient spatial precision, while DINOv3 provides better spatial features but transfers less effectively to the LLM. C-RADIOv4-H achieves the strongest overall performance among the evaluated vision encoders. 
More critically, with T-REN, regions lose spatial connectivity during VLM training. Even with C-RADIOv4-H, after just 1,000 training steps, ${\sim}$37\% of region tokens become spatially disconnected in text-heavy images from PixMo-Cap. A single position ID cannot faithfully represent multiple disconnected components, making it difficult for the model to associate visual content with its location. This is particularly problematic for structured text images (Figure~\ref{fig:disjoint-regions} in Appendix). Enforcing connectivity produces near-rectangular regions, sacrificing the content adaptivity that motivates region tokenization. Freezing T-REN alleviates disjoint regions but degrades performance.

\input{tables/region-generation-methods}

\boldheader{Feature-based clustering}
Following DiVT~\citep{DiVT}, we explore constructing regions by grouping patches with similar ViT features. We first adopt DiVT's training setup, keeping the vision encoder frozen while training the connector and fine-tuning the LLM. However, freezing the ViT limits downstream performance compared to end-to-end training. We therefore allow the ViT to train jointly with the LLM and connector, but find that as patch embeddings adapt, the resulting regions become spatially disconnected, reproducing the failure observed with T-REN. To avoid this trade-off, we decouple region generation from VLM training by using a separate frozen ViT for patch grouping while allowing the VLM's vision encoder to train freely. With C-RADIOv4-H as the frozen encoder, this variant approaches the performance of the fully trained patch baseline while using fewer visual tokens. However, it requires a second ViT forward pass for every image, adding parameters, latency, and memory overhead.

\boldheader{Learned segmenter}
Segmentation models like SAM~\citep{SAM} can be used to obtain very precise segmentations. However, generating all regions for an image with SAM is slow and computationally expensive. More importantly, SAM does not guarantee complete image coverage, leaving some patches without a region assignment (Figure~\ref{fig:region-generation-methods} and App.~\ref{app:unseen-domains}). This is particularly problematic for pointing, where locations excluded from the visual representation cannot be reliably predicted. EPOC~\citep{EPOC} addresses incomplete coverage using a learned boundary detector followed by watershed segmentation. However, its region construction introduces preprocessing overhead. For instance, with eight dataloader workers, FH achieves a segmentation throughput of 159 images/s, compared to 23 images/s for EPOC, a ${\sim}7{\times}$ difference. FH preprocessing can be hidden behind VLM execution, whereas EPOC's accelerator-dependent pipeline limits this overlap. We also find that EPOC underperforms FH on text-heavy images, which account for much of the PixMo-Cap performance gap in Table~\ref{tab:region-generation-methods}.

\boldheader{Training-free pixel clustering}
Methods such as SLIC~\citep{SLIC} and FH~\citep{FH} avoid the limitations of the previous approaches. They are training-free and operate directly on image pixels, making region quality independent of the VLM's vision encoder. They also provide spatially connected regions and complete image coverage without adding model parameters or forward passes. Although these methods tend to produce finer partitions and achieve less token reduction than T-REN or SAM, they still provide content-adaptive visual tokens while satisfying the requirements for end-to-end VLM training and spatial grounding. \rone{Computing region assignments from the input alone also simplifies training. Because they do not depend on the model, both the number and spatial extent of region tokens are known before the image enters the model, so dataloader workers can assign position IDs and pack variable-length sequences in parallel with GPU computation. In contrast, methods whose regions come from a network must either run that network during data loading, adding preprocessing cost (EPOC: 23 vs.\ 159 images/s for FH), or, when the network is trained with the VLM (T-REN, DiVT with a trainable ViT), defer region construction to the forward pass.} The latter requires placeholder sequences with a fixed visual token budget, wasting computation when the budget is too large and truncating region sequences when it is too small. 

\ronetwo{\boldheader{Summary}
Three properties distinguish region strategies that work across tasks from those that do not. (1) Regions must cover the full image, which SAM does not guarantee. (2) They must stay spatially connected during end-to-end training, which regions derived from the VLM's own features do not. (3) They must also preserve fine structure such as small text, which SLIC and most learned methods handle poorly. Further, semantic grouping adds little, because the LLM can aggregate information across object parts just as it does across patches. FH satisfies all three properties without adding parameters or forward passes, so we use it in \rvlm{}.}

%% file: tables/task-comparison.tex
\begin{table}[t]
    \caption{{\bf Comparison with open-weight VLMs.} Our identically trained patch and region models achieve competitive performance across image QA, pointing, and counting benchmarks. Thus, \pvlm{} provides a strong baseline for isolating the effect of region-based visual tokenization. For TextVQA, we report validation-set results for our models because the test server is unavailable.}
    \vspace{-0.7em}
    \label{tab:task-comparison}
    \centering
    \small
    \setlength{\tabcolsep}{4pt}
    \renewcommand{\arraystretch}{1.15}
    \begin{tabular}{l ccccccc}
        \toprule
        \textbf{Model}
            & \textbf{VQAv2}
            & \textbf{AI2D}
            & \textbf{PixMo-Point}
            & \textbf{PixMo-Count}
            & \textbf{CountBench}
            & \textbf{TextVQA}
            & \textbf{DocVQA} \\
        \midrule
        InternVL3.5-4B
            & 78.1
            & 82.6
            & --
            & 62.4
            & 82.2
            & 77.9
            & 92.4 \\
        PLM-3B
            & 84.4
            & 90.9
            & --
            & 63.0
            & 87.1
            & 84.3
            & 93.8 \\
        Qwen3-VL-4B
            & 81.7
            & 84.1
            & 43.7
            & 58.0
            & 85.5
            & 81.0
            & \textbf{95.3} \\
        Molmo2-4B
            & 86.6
            & \textbf{95.6}
            & 83.4
            & 88.1
            & \textbf{93.9}
            & \underline{85.0}
            & 92.4 \\
        \midrule
        \pvlm{}-4B
            & \textbf{87.1}
            & \underline{95.4}
            & \underline{85.2}
            & \underline{90.1}
            & 90.2
            & \textbf{85.5}
            & \underline{94.4} \\
        \rowcolor{blue} \rvlm{}-4B
            & \underline{87.0}
            & 95.1
            & \textbf{86.4}
            & \textbf{90.4}
            & \underline{91.8}
            & \underline{85.0}
            & 93.7 \\
        \bottomrule
    \end{tabular}
\end{table}

%% file: tables/native-res-overall-results.tex
\begin{table*}[t]
    \caption{\textbf{Patch vs.\ region tokenization.} \rvlm{} preserves or improves overall performance over the identically trained patch baseline while using fewer visual tokens. Its main advantage, however, comes from enabling higher-resolution processing, as shown in Figure~\ref{fig:fixed-tok}. Example-weighted averages give equal weight to every example, while family-weighted averages give equal weight to each task family. Token counts are averaged across examples.}
    \vspace{-0.7em}
    \label{tab:native-res-overall-results}
    \centering
    \small
    \begin{tabular}{l ccc}
        \toprule
        \textbf{Model} & \textbf{Example-weighted avg.} & \textbf{Family-balanced avg.} & \textbf{Token count} \\
        \midrule
        \pvlm{} & 82.8 & 79.1 & 985 \\
        \rowcolor{blue} \rvlm{} & 
            83.4 \textcolor{green}{\scriptsize\bfseries (\(\uparrow\)0.6)} & 
            79.8 \textcolor{green}{\scriptsize\bfseries (\(\uparrow\)0.7)} & 
            773 \textcolor{green}{\scriptsize\bfseries (\(\downarrow\)21.5\%)} \\
        \bottomrule
    \end{tabular}

    \vspace{2em}

    \begin{subtable}[t]{0.32\linewidth}
        \centering
        \caption{{\bf Reasoning / understanding.} On general Image QA, regions remain near parity with patches. On document-style VQA they are competitive on ChartQA, DocVQA, and TextVQA, with a clear drop only on InfoVQA.}
        \vspace{-0.5em}
        \label{tab:qa}
        \scriptsize
        \begin{tabular}{l cc}
            \toprule
            \textbf{Task} & \textbf{\pvlm{}} & \textbf{\rvlm{}} \\
            \midrule
            \rowcolor{black!8} \multicolumn{3}{c}{\textbf{Image QA}} \\
            A-OKVQA DA & 75.5 & 74.8 \\
            A-OKVQA MC & 88.6 & 88.2 \\
            AI2D & 95.4 & 95.1 \\
            OK-VQA & 65.8 & 66.0 \\
            ScienceQA & 98.2 & 97.8 \\
            VQAv2 & 87.2 & 87.1 \\
            ChartQA & 83.7 & 82.9 \\
            DocVQA & 93.2 & 92.8 \\
            InfoVQA & 79.2 & 74.3 \\
            TextVQA & 85.5 & 85.0 \\
            \rowcolor{black!4} \textbf{Task average} & 85.2 & 84.4 \\
            \rowcolor{black!4} \tval{\textbf{Token count}} & \tval{793} & \tval{700} \\
            \bottomrule
        \end{tabular}
    \end{subtable}
    \hfill
    \begin{subtable}[t]{0.32\linewidth}
        \centering
        \caption{{\bf Grounding.} Regions match or beat patches in every family while using 32-41\% fewer visual tokens.}
        \vspace{-0.5em}
        \label{tab:grounding}
        \scriptsize
        \begin{tabular}{l cc}
            \toprule
            \textbf{Task} & \textbf{\pvlm{}} & \textbf{\rvlm{}} \\
            \midrule
            \rowcolor{black!8} \multicolumn{3}{c}{\textbf{Pointing}} \\
            OSWorld-G & 51.6 & 52.0 \\
            PixMo-Point & 85.2 & 86.4 \\
            PointBench & 63.8 & 63.3 \\
            SA-Co Gold & 87.5 & 90.1 \\
            ScreenSpot Pro & 29.3 & 26.0 \\
            ScreenSpot v2 & 82.7 & 85.0 \\
            \rowcolor{black!4} \textbf{Task average} & 66.7 & 67.1 \\
            \rowcolor{black!4} \tval{\textbf{Token count}} & \tval{1,793} & \tval{1,065} \\
            \midrule
            \rowcolor{black!8} \multicolumn{3}{c}{\textbf{Counting}} \\
            CountBenchQA & 90.2 & 91.8 \\
            PixMo-Count & 90.1 & 90.4 \\
            \rowcolor{black!4} \textbf{Task average} & 90.2 & 91.1 \\
            \rowcolor{black!4} \tval{\textbf{Token count}} & \tval{1,655} & \tval{990} \\
            \midrule
            \rowcolor{black!8} \multicolumn{3}{c}{\textbf{Segmentation}} \\
            PixMo-Seg & 58.0 & 59.8 \\
            RefCOCO & 82.0 & 82.5 \\
            RefCOCO+ & 78.5 & 78.7 \\
            RefCOCOg & 80.3 & 81.1 \\
            SA-Co Gold & 50.7 & 54.2 \\
            \rowcolor{black!4} \textbf{Task average} & 69.9 & 71.2 \\
            \rowcolor{black!4} \tval{\textbf{Token count}} & \tval{1,390} & \tval{950} \\
            \bottomrule
        \end{tabular}
    \end{subtable}
    \hfill
    \begin{subtable}[t]{0.32\linewidth}
        \centering
        \caption{{\bf 3D.} We report scale-invariant $\delta_{1.25}$ for point-to-depth across 12 datasets. Regions match or beat patches on average while using 43\% fewer visual tokens.}
        \vspace{-0.5em}
        \label{tab:3d}
        \scriptsize
        \begin{tabular}{l cc}
            \toprule
            \textbf{Task} & \textbf{\pvlm{}} & \textbf{\rvlm{}} \\
            \midrule
            \rowcolor{black!8} \multicolumn{3}{c}{\textbf{Depth}} \\
            Argoverse 2 & 77.9 & 81.8 \\
            Aria AEA & 68.5 & 69.5 \\
            Aria Nymeria & 54.6 & 55.6 \\
            DDAD & 81.7 & 87.7 \\
            DIODE & 92.2 & 94.0 \\
            EgoExo4D Ego & 89.1 & 88.9 \\
            EgoExo4D Exo & 84.0 & 82.9 \\
            nuScenes & 84.7 & 88.8 \\
            NYU Depth v2 & 97.8 & 97.5 \\
            Stereo4D & 89.0 & 88.6 \\
            SUN RGB-D & 98.2 & 98.2 \\
            Waymo & 84.1 & 90.8 \\
            \rowcolor{black!4} \textbf{Task average} & 83.5 & 85.4 \\
            \rowcolor{black!4} \tval{\textbf{Token count}} & \tval{1,617} & \tval{916} \\
            \bottomrule
        \end{tabular}
    \end{subtable}
    \vspace{-1em}
\end{table*}

%% file: tables/inference-cost.tex
\begin{table}[t]
    \caption{\textbf{Inference cost.} Throughput is examples processed per second of wall-clock time over 512 randomly sampled examples from PixMo-Cap, including data loading and region construction. Generation is capped at one output token, so throughput reflects prefill. Measurements use a single NVIDIA H100 after 64 warm-up batches. We report the mean and standard deviation over five runs. Shorter visual sequences increase throughput and reduce peak memory, with larger gains at larger batch sizes. Table~\ref{tab:inference-cost-sweep} shows a finer-grained cost comparison across image resolutions.}
    \vspace{-0.7em}
    \label{tab:inference-cost}
    \centering
    \small
    \setlength{\tabcolsep}{5pt}
    \renewcommand{\arraystretch}{1.2}
    \begin{tabular}{l cc cc}
        \toprule
        \textbf{Model}
        & \multicolumn{2}{c}{\textbf{Batch size 1}}
        & \multicolumn{2}{c}{\textbf{Batch size 8}} \\
        \cmidrule(lr){2-3}\cmidrule(lr){4-5}
        & \textbf{Throughput (ex/s)} $\uparrow$
        & \textbf{Peak mem.\ (GiB)} $\downarrow$
        & \textbf{Throughput (ex/s)} $\uparrow$
        & \textbf{Peak mem.\ (GiB)} $\downarrow$ \\
        \midrule
        \pvlm{} & $13.39{\pm}0.16$ & $19.98$ & $19.27{\pm}0.07$ & $26.75$ \\
        \rowcolor{blue} \textbf{\rvlm{}} & $\mathbf{15.90{\pm}0.38}$ & $\mathbf{19.75}$ & $\mathbf{26.18{\pm}0.07}$ & $\mathbf{24.69}$ \\
        $\Delta$ & \textcolor{green}{$\mathbf{+18.7\%}$} & \textcolor{green}{$\mathbf{-1.2\%}$} & \textcolor{green}{$\mathbf{+35.9\%}$} & \textcolor{green}{$\mathbf{-7.7\%}$} \\
        \bottomrule
    \end{tabular}
\end{table}

%% file: tables/region-generation-methods.tex
\begin{table*}[t]
    \caption{{\bf Comparing region generation strategies.} Properties are derived from the failure modes discussed in Section~\ref{sec:constructing-regions}. Performance is reported on PixMo-Cap (Cap) and PixMo-Point (Point) after stage-1 training with Qwen3-1.7B as the language model.}
    \vspace{-0.7em}
    \label{tab:region-generation-methods}
    \centering
    \small
    \setlength{\tabcolsep}{2.8pt}
    \renewcommand{\arraystretch}{1.15}
    \begin{tabular}{lccccccc}
        \toprule
        \textbf{Strategy}
        & \makecell{\textbf{Complete}\\\textbf{patch}\\\textbf{coverage}}
        & \makecell{\textbf{Connectivity}\\\textbf{preserved after}\\\textbf{training}}
        & \makecell{\textbf{No external}\\\textbf{region}\\\textbf{encoder}}
        & \makecell{\textbf{Encoder-}\\\textbf{independent}\\\textbf{regions}}
        & \makecell{\textbf{Regions}\\\textbf{at image}\\\textbf{loading}}
        & \makecell{\textbf{Cap}\\\textbf{$\uparrow$}}
        & \makecell{\textbf{Point}\\\textbf{$\uparrow$}} \\
        \midrule
        \rowcolor{black!8} \multicolumn{8}{c}{\textbf{Region encoder network}} \\
        T-REN w/ SigLIP-2 & \cmark & \xmark & \cmark & \xmark & \xmark & 52.9 & 77.6 \\
        T-REN w/ DINOv3 & \cmark & \xmark & \cmark & \xmark & \xmark & 51.9 & 80.8 \\
        T-REN w/ C-RADIOv4-H & \cmark & \xmark & \cmark & \xmark & \xmark & 52.9 & 84.4 \\
        \rowcolor{black!8} \multicolumn{8}{c}{\textbf{Feature-based clustering}} \\
        DiVT (Frozen ViT) & \cmark & \cmark & \cmark & \xmark & \xmark & 51.4 & 79.8 \\
        DiVT (Trainable ViT) & \cmark & \xmark & \cmark & \xmark & \xmark & 52.6 & 80.2 \\
        DiVT (Dual ViT) & \cmark & \cmark & \xmark & \xmark & \xmark & 53.1 & 86.4 \\
        \rowcolor{black!8} \multicolumn{8}{c}{\textbf{Learned segmenter}} \\
        SAM & \xmark & \cmark & \xmark & \cmark & \xmark & -- & -- \\
        EPOC & \cmark & \cmark & \xmark & \cmark & \xmark & 53.1 & 86.6 \\
        \rowcolor{black!8} \multicolumn{8}{c}{\textbf{Training-free pixel clustering}} \\
        SLIC & \cmark & \cmark & \cmark & \cmark & \cmark & 53.0 & 86.5 \\
        \rowcolor{blue} FH (ours) & \cmark & \cmark & \cmark & \cmark & \cmark & 53.8 & 86.6 \\
        \bottomrule
    \end{tabular}
    \vspace{-1em}
\end{table*}

%% file: sections/5-conclusion.tex
\section{Conclusion}
\label{sec:conclusion}
We present \rvlm{}, a VLM that replaces fixed spatial pooling with content-adaptive region tokens. \ronetwo{Our analysis of region construction shows that, for general-purpose VLMs, regions must cover the full image, stay spatially connected during end-to-end training, and preserve fine structure such as small text, while semantic grouping adds little. A fast, training-free pixel-based segmenter therefore suffices.} Region tokens preserve or improve performance while using fewer visual tokens, and their benefits grow with image resolution. By retaining complete patch coverage and explicit patch-to-region correspondence, \rvlm{} supports precise grounded prediction without adding model parameters. \rtwo{These results hold on unseen pathology and satellite images, at the 1.7B scale, and on video without video training.} Together, these findings establish region-based tokenization as an efficient alternative to fixed spatial pooling for general-purpose VLMs. Future work can extend this approach to \rtwo{long} video\rtwo{, for example by grouping regions across frames, and evaluate it with sparse mixture-of-experts backbones}.

\subsection*{AI Use Statement}
In this work, we used generative AI tools to improve the clarity and readability of author-written text, assist with manuscript formatting, create exploratory visualizations used only during development, and refactor and optimize existing code without changing the underlying methods. The authors manually reviewed all AI-assisted material and inspected AI-generated code. Code changes were validated by confirming that their outputs matched those of the original implementations. We also verified factual claims and citations against original sources and checked table entries against source data. We take responsibility for the final content of this work, including all text, claims, code, and artifacts produced with the aid of generative AI.

\subsection*{Reproducibility Statement}
We provide the model architecture, region construction procedure, patch-to-region assignment, and spatial grounding formulation in Section~\ref{sec:method}. Appendix~\ref{app:implementation-details} documents the training configuration, data mixtures, preprocessing parameters, task formulations, evaluation datasets, and metrics. Section~\ref{sec:experiments} and Appendix~\ref{app:implementation-details} describe the experimental setups for comparisons with patch-based models, token compression methods, and alternative region construction strategies. 

%% file: sections/6-appendix.tex
\clearpage
\appendix
\section*{Appendix}

We provide training and evaluation details in Section~A. Section~B reports additional experiments on FH parameters, connector width, and resolution scaling\ronetwo{, along with design ablations for pointing with region tokens, generalization to unseen domains, alternative position encoding, model scale, and text-only and video evaluation}. Section~C discusses limitations.

\section{Implementation Details}
\label{app:implementation-details}

\subsection{Architecture}
We adopt MolmoPoint~\citep{MolmoPoint} as the basis for our patch baseline (\pvlm{}) and the starting architecture for \rvlm{}, motivated by its strong performance among similarly sized open-weight VLMs (Table~\ref{tab:task-comparison}). We initialize the LLM from Qwen3-4B-Instruct~\citep{Qwen3} and the ViT from SigLIP2-SO400M/14~\citep{SigLIP2}. The LLM has a hidden dimension of $2560$, $36$ layers, $32$ attention heads, and $8$ KV heads. The vision encoder processes $378$-pixel inputs with a patch size of $14$. The connector concatenates ViT features from layers $-3$ and $-9$, then applies 16-head cross-attention using the mean of each patch group as the query and its constituent features as keys and values. The pooled representation is projected to the LLM embedding space through an MLP. We use adaptive overlapping crops with crop margins of $[4,4]$.

\subsection{Training recipe}
We train \rvlm{} in two stages, jointly optimizing the vision encoder, connector, and LLM throughout. Stage 1 performs multimodal pretraining for $31{,}971$ steps. Stage 2 initializes from the Stage-1 checkpoint and performs supervised fine-tuning for another $24{,}000$ steps. Table~\ref{tab:training-hparams} summarizes the training configuration, and Table~\ref{tab:training-mixtures} shows the data mixtures. The training recipe is adopted from concurrent work in preparation and is not a contribution of this paper.

\subsection{Task Formulation}
\label{app:task-formulation}
We summarize how the model performs each task. The pointing formulation follows MolmoPoint~\citep{MolmoPoint}, while the segmentation and point-to-depth formulations are adopted from concurrent work in preparation and are not contributions of this paper.

\boldheader{Image QA and captioning}
We formulate image question answering and captioning as autoregressive text generation. Given an image and a textual prompt, the model conditions on the region-token representation of the image and generates an answer or caption using a standard next-token objective. No task-specific prediction head or output representation is used for these tasks.

\boldheader{Pointing and counting}
We represent an image location using a coarse-to-fine hierarchy. For each point, the model first selects a region token, then a ViT patch assigned to that region, and finally one of $4{\times}4$ spatial cells within the selected patch. \rone{Region selection scores the hidden state of the first pointing token generated by the LLM against keys from the LLM hidden states of the region tokens; patch selection scores the second against keys from the retained ViT patch features of the patches assigned to the selected region token; finally, the third predicts the bin within the selected patch using a linear classifier.} The center of the predicted bin is mapped back to image coordinates using the retained patch-to-region assignment. These decisions are generated autoregressively using dedicated classification heads, with cross-entropy supervision at each level. A learned end-of-points class terminates the point sequence. For counting, the model predicts one point for each instance that matches the query using the same grounding representation. It subsequently generates the cardinality of the point set as text. Thus, counting is grounded in an explicit set of localized instances rather than being performed solely through direct scalar regression.

\boldheader{Segmentation}
For referring segmentation, the model first selects an anchor region for the target mask, supervised using the regions with maximal target-mask coverage. \rone{Conditioned on this anchor, it then generates a special segmentation token whose hidden state serves as a query. Region logits are its dot product with the region-token keys. Patch logits are its dot product with each retained ViT patch feature after a linear projection. Cell logits are its dot product with 16 cell features per patch, produced by a linear layer that maps each ViT patch feature to one feature per $4{\times}4$ cell; no additional feature map is used.}
The three scores are summed for each bin and mapped back to the image using the retained patch-to-region assignment. The resulting mask probabilities are bilinearly resized to the original image resolution and thresholded at $0.5$. Multiple masks are generated by repeating the anchor-and-mask prediction sequence. Training uses region-selection supervision together with focal and Dice losses for the dense masks.

\boldheader{Point-to-depth}
Given an image and one or more queried locations, the model predicts the depth at each point. Query points are encoded in the prompt, and the output contains a corresponding coarse-to-fine point representation for each query, followed by a measurability indicator and, when measurable, its depth in meters. Depth is generated as text to a precision of two decimal places.

\subsection{Evaluation}
We evaluate on 35 tasks spanning image QA (including document/OCR QA), pointing, counting, referring segmentation, and point-to-depth. Table~\ref{tab:eval-tasks} lists each dataset, the metric reported in the results, and a one-line task description. Unless noted, scores are percentages, with higher scores indicating better performance.

\input{tables/training-hparams}
\input{tables/training-mixtures}
\input{tables/eval-tasks}

\section{Additional Experiments}
\label{app:additional-ablations}

\subsection{Ablating FH Segmentation Parameters}
\label{app:fh-ablation}

We study how the scale ($k$), the minimum region size ($\mathrm{min\_region\_size}$), and the working resolution ($\mathrm{max\_side}$) of Felzenszwalb-Huttenlocher (FH) hyperparameters affect downstream accuracy and visual token count. All runs use the same \rvlm{} SFT checkpoint, only inference-time FH settings change. We evaluate a six-task subset spanning general VQA, document OCR, pointing, counting, referring segmentation, and depth. The results are shown in Table~\ref{tab:fh-ablation}.

\input{tables/fh-ablations}

\boldheader{General VQA and depth are robust}
A-OKVQA DA stays within $74.8$-$75.5$ and NYU $\delta_{1.25}$ within 
$97.4$-$97.6$ across all settings. For these tasks, FH mainly changes token compression, not accuracy.

\boldheader{Text-rich understanding benefits from finer segmentation}
Coarser segmentation hurts document OCR. Raising $k$ from $100$ to $400$ drops InfoVQA by $2.2$ points, and raising $\mathrm{min\_size}$ from $32$ to $128$ drops it by $3.6$. Conversely, a higher FH working resolution helps, with score reaching $75.2$ at $\mathrm{max\_side}{=}1024$.

\boldheader{Grounding degrades gently under coarsening}
PixMo-Point, PixMo-Count, and RefCOCO typically move by ${\le}1.5$ points when regions become coarser or the FH working resolution reduces. Aggressive compression ($k{=}400$ or $\mathrm{max\_side}{=}384$) trades a small grounding loss for substantially fewer tokens.

\boldheader{Tokens track FH granularity}
Mean visual tokens fall from $1171$ at $k{=}50$ to $686$ at $k{=}400$, and from $1032$ at $\mathrm{min\_size}{=}16$ to $736$ at $128$. Increasing $\mathrm{max\_side}$ raises tokens mainly on high-resolution OCR/pointing images (InfoVQA: $1053{\to}1669$ from $384{\to}1024$), while smaller natural images saturate once $\mathrm{max\_side}$ exceeds their native size.

\rtwo{\textbf{Practical heuristics.}
For our evaluations, we use a single FH setting for all tasks ($k{=}100$, minimum region size 32, maximum side 768; Sec.~\ref{sec:modeling}). We recommend lowering $k$ or the minimum region size for text-heavy inputs, and raising them when the token budget matters more than fine detail. The working resolution matters mainly for small text (InfoVQA: 71.0 at maximum side 384 vs.\ 75.2 at 1024).}

\subsection{Connector Width Ablation}
\label{app:connector-width-ablation}
The region connector aggregates at most $w$ ViT patches into each region token. Regions containing more than $w$ patches are split into multiple connector rows, while shorter rows are padded to width $w$. Increasing $w$ can therefore reduce the number of visual tokens passed to the LLM, but also increases the size of the padded connector input. We analyze this structural trade-off to motivate our default choice of $w=16$. Specifically, we sweep $w\in\{4,8,16,32,64,128\}$ while keeping the FH configuration fixed at $(k,\mathrm{min\_size},\mathrm{max\_side})=(100,32,768)$. Our stratified sample contains $1{,}480$ images from A-OKVQA DA, InfoVQA, PixMo-Point, PixMo-Count, RefCOCO, and NYU Depth v2. Table~\ref{tab:connector-width-ablation} summarizes our structural analysis.

\boldheader{\boldmath Most regions fit within $w{=}16$}
At $w{=}4$, $31.1\%$ of FH regions require splitting, compared to only $7.8\%$ at $w{=}16$ (Split in Table~\ref{tab:connector-width-ablation}). Thus, the default width accommodates $92.2\%$ of regions without splitting, leaving limited room for further reductions in visual tokens.

\boldheader{Wider connectors introduce substantial padding}
The additional token savings come at the cost of a larger padded connector input. Increasing $w$ from $16$ to $128$ reduces visual tokens by only $19.3\%$, while increasing the padding fraction from $62.2\%$ to $94.1\%$. Consequently, the estimated bf16 connector input memory grows from $68.9$ to $444.8$ MiB, a $6.5\times$ increase. This trade-off motivates $w=16$ as our default connector width.

\boldheader{\boldmath Visual-token reduction saturates near $w{=}16$}
Table~\ref{tab:connector-width-ablation} shows that increasing $w$ from $4$ to $16$ reduces the mean visual-token count from $1{,}814$ to $980$, a $46.0\%$ reduction. This captures $81.5\%$ of the total token reduction observed between $w=4$ and $w=128$. Increasing the width beyond $16$ yields diminishing returns.

\input{tables/connector-width-ablation}

\subsection{Resolution Scaling}
\label{app:resolution-scaling}

\boldheader{Performance across image resolutions}
Table~\ref{tab:native-res-bucketted-results} breaks down the native-resolution evaluation into five image-size ranges. At lower resolutions, \rvlm{} and \pvlm{} perform comparably, but the gap widens as image size increases. In the smallest range (0.14-0.41 MP), both models achieve an example-weighted score of $85.3$, with \rvlm{} using 4.5\% fewer visual tokens. In the largest range (2.08-2.63 MP), \rvlm{} improves performance by 5.3 points while reducing the average token count by 51.8\%. The family-balanced gain also grows from 0.3 to 9.0 points across these ranges.

The advantage is particularly pronounced on spatial tasks. In the largest resolution range, \rvlm{} improves the task averages by 5.6 points on pointing, 25.9 points on counting, 8.7 points on segmentation, and 5.5 points on depth estimation. Image QA remains broadly comparable. These results suggest that region tokens make better use of high-resolution visual information, particularly when predictions depend on fine spatial details. Since each resolution range contains a different set of task examples, absolute scores should not be compared directly across ranges; the relevant comparison is between \pvlm{} and \rvlm{} within each range.

\boldheader{Inference cost across image resolutions}
Table~\ref{tab:inference-cost-sweep} measures inference cost as the number of high-resolution crops increases. At one and four crops, \pvlm{} is slightly faster at batch size 1, where the savings from shorter visual sequences do not yet offset the overhead of region pooling. At batch size 8, the models have nearly identical latency at four crops. From eight crops onward, \rvlm{} achieves lower latency and peak memory at both batch sizes.

The efficiency gap continues to widen with resolution. At 32 crops and batch size 1, \rvlm{} reduces generation latency from $162.1$ to $105.0$ ms. At batch size 8, latency falls from $144.6$ to $83.7$ ms, while peak GPU memory decreases from $39.57$ to $30.15$ GiB. These trends reflect the growing benefit of reducing the visual sequence processed by the LLM as image resolution increases, even though the vision encoder still processes the full patch grid.

\input{tables/native-res-bucketted-results}
\input{tables/inference-cost-sweep}



\subsection{\texorpdfstring{\ronetwo{What Region Tokens Need for Grounding}}{What Region Tokens Need for Grounding}}
\label{app:dense-outputs}
\rtwo{To support precise spatial prediction, region tokens must satisfy two properties. Every patch must belong to exactly one region, so that every target is reachable and unambiguous, and no region may contain more patches than the decoder can choose among, so that every patch can be selected. We test these properties in 1.7B development runs, replacing patch tokens with region tokens under two pointing formulations (Table~\ref{tab:dense-outputs}).}
\ronetwo{With direct coordinate output, regions trail patches by 4.8 points, because the LLM must regress precise locations from tokens of varying spatial extent.}
\rtwo{Coarse-to-fine decoding removes this burden, but with unmodified regions it scores only 10.0, since those regions violate both properties. Some patches belong to no region or to several, and the second stage considers a fixed number of patches per region, so patches in very large regions cannot be selected. Assigning every patch to exactly one region and splitting large regions into contiguous groups (Sec.~\ref{sec:modeling}) restores both properties, and regions then surpass patches (86.6 vs.\ 84.4).}

\begin{table}[t]
    \centering
    \small
    \caption{\ronetwo{{\bf Demonstrating importance of region splitting and ensuring exhaustive, exclusive patch assignment.} Naively substituting patch tokens with region tokens fails catastrophically on PixMo-Point (row 2). Ensuring region tokens are capped to span a maximum number of patches and the partitioning is exhaustive and exclusive (row 3) enables RVLM to outperform patch-based baseline on pointing. The results are reported at 1.7B scale.}}
    \vspace{-0.6em}
    \label{tab:dense-outputs}
    \begin{tabular}{lcc}
    \toprule
    \textbf{Representation unit} & \textbf{Direct coordinate output} & \textbf{Coarse-to-fine prediction} \\
    \midrule
    Patches & 77.5 & 84.4 \\
    Regions (naive) & 72.7 & 10.0 \\
    Regions (\rvlm{}) & n/a & 86.6 \\
    \bottomrule
    \end{tabular}
\end{table}

\begin{table}[t]
    \centering
    \small
    \caption{\rtwo{{\bf Performance on unseen domains.} RVLM performs on par with PVLM on pathology and satellite images, which are unseen during training. The reported token counts are mean visual tokens per image (\pvlm{} $\rightarrow$ \rvlm{}).}}
    \vspace{-0.6em}
    \label{tab:unseen-domains}
    \begin{tabular}{l ccc}
    \toprule
    \textbf{Benchmark (metric)} & \textbf{\pvlm{}} & \textbf{\rvlm{}} & \textbf{Token count} \\
    \midrule
    \rowcolor{black!8} \multicolumn{4}{c}{\textbf{Pathology}} \\
    \midrule
    PathMMU (accuracy) & 45.3 & 47.0 & 1842 $\rightarrow$ 972 \\
    PathVQA (accuracy) & 63.1 & 63.7 & 1850 $\rightarrow$ 869 \\
    \midrule
    \rowcolor{black!8} \multicolumn{4}{c}{\textbf{Satellite imagery}} \\
    \midrule
    RSVQA-HR (accuracy) & 47.9 & 47.3 & 1829 $\rightarrow$ 853 \\
    LoveDA (segmentation cIoU) & 62.6 & 63.4 & 1829 $\rightarrow$ 821 \\
    \bottomrule
    \end{tabular}
\end{table}

\begin{table}[t]
    \centering
    \small
    \caption{\rtwo{{\bf FH segments unseen domains more reliably than SAM.} FH achieves higher boundary recall (BR) on every dataset and covers every pixel by construction, whereas SAM leaves 13--29\% of pixels uncovered.}}
    \vspace{-0.6em}
    \label{tab:boundary-recall}
    \begin{tabular}{lcccc}
    \toprule
    Dataset & SAM BR & FH BR & SAM coverage & FH coverage \\
    \midrule
    LoveDA & 39.0 & 54.8 & 71.0\% & 100\% \\
    Total-Text (words) & 34.2 & 64.2 & 87.1\% & 100\% \\
    Total-Text (strokes) & 60.7 & 83.5 & 87.1\% & 100\% \\
    \bottomrule
    \end{tabular}
\end{table}

\begin{table}[t]
    \centering
    \small
    \caption{\rtwo{{\bf Comparing position encoding for region tokens.} The reported results are for \rvlm{}-1.7B after stage-1 training.}}
    \vspace{-0.6em}
    \label{tab:position-encoding}
    \begin{tabular}{lcccc}
    \toprule
    \textbf{Position encoding} & \textbf{Added params} & \textbf{PixMo-Cap} & \textbf{PixMo-Point} & \textbf{PixMo-Count} \\
    \midrule
    Region-mask absolute PE + 1D RoPE & Yes & 53.6 & 86.7 & 84.7 \\
    3D M-RoPE (ours) & No & 53.8 & 86.6 & 84.4 \\
    \bottomrule
    \end{tabular}
\end{table}

\begin{table}[t]
    \centering
    \small
    \caption{\rtwo{\bf Stage-1 results at 1.7B.}}
    \vspace{-0.6em}
    \label{tab:scale-stage1}
    \begin{tabular}{lccc}
    \toprule
    \textbf{Model} & \textbf{PixMo-Cap} & \textbf{PixMo-Point} & \textbf{PixMo-Count} \\
    \midrule
    \pvlm{}-1.7B & 54.1 & 84.4 & 84.6 \\
    \rvlm{}-1.7B & 53.8 & 86.6 & 84.4 \\
    \bottomrule
    \end{tabular}
\end{table}

\begin{table}[t]
    \centering
    \small
    \caption{\rtwo{\bf Stage-2 results at 1.7B.}}
    \vspace{-0.6em}
    \label{tab:scale-stage2}
    \begin{tabular}{lcccccc}
    \toprule
    \textbf{Model} & \textbf{COCO} & \textbf{\makecell{A-OKVQA\\(MC)}} & \textbf{\makecell{A-OKVQA\\(DA)}} & \textbf{InfoVQA} & \textbf{ChartQA} & \textbf{DocVQA} \\
    \midrule
    \pvlm{}-1.7B & 80.5 & 79.4 & 62.5 & 56.3 & 77.6 & 85.5 \\
    \rvlm{}-1.7B & 81.5 & 80.4 & 63.2 & 57.3 & 77.0 & 84.4 \\
    \bottomrule
    \end{tabular}
\end{table}

\begin{table}[t]
    \centering
    \small
    \caption{\rtwo{\bf Performance on text-only benchmarks.}}
    \vspace{-0.6em}
    \label{tab:text-only}
    \begin{tabular}{lccc}
    \toprule
    \textbf{Model} & \textbf{MMLU} & \textbf{ARC-C} & \textbf{GSM8K} \\
    \midrule
    \pvlm{}-4B & 67.7 & 87.6 & 89.0 \\
    \rvlm{}-4B & 67.8 & 87.0 & 88.9 \\
    \bottomrule
    \end{tabular}
\end{table}

\begin{table}[t]
    \centering
    \small
    \caption{\rtwo{{\bf Performance on video benchmarks}. \rvlm{} uses 3$\times$ fewer visual tokens compared to PVLM. Neither model is trained on videos.}}
    \vspace{-0.6em}
    \label{tab:video}
    \begin{tabular}{lccc}
    \toprule
    \textbf{Model} & \textbf{MVBench} & \textbf{MLVU} & \textbf{NeXT-QA} \\
    \midrule
    \pvlm{}-4B & 50.9 & 44.0 & 69.9 \\
    \rvlm{}-4B & 50.6 & 44.2 & 69.4 \\
    \bottomrule
    \end{tabular}
\end{table}

\subsection{\texorpdfstring{\rtwo{Generalization to Unseen Domains}}{Generalization to Unseen Domains}}
\label{app:unseen-domains}
\rtwo{We evaluate \pvlm{} and \rvlm{} on pathology and satellite benchmarks, on which neither model is trained (Table~\ref{tab:unseen-domains}). Because the two models differ only in tokenization, this directly tests whether FH regions transfer. \rvlm{} improves on PathMMU (+1.7), PathVQA (+0.6), and LoveDA (+0.8) and trails slightly on RSVQA-HR ($-$0.6), while using about half the visual tokens. We also measure boundary recall (BR) against ground-truth masks for SAM ViT-H and for FH at \rvlm{}'s settings (Table~\ref{tab:boundary-recall}). FH achieves higher BR on every out-of-domain set and covers every pixel by construction, whereas SAM leaves 13--29\% of pixels uncovered.}

\subsection{\texorpdfstring{\rtwo{Alternative Position Encoding}}{Alternative Position Encoding}}
\label{app:position-encoding}
\rtwo{A region token's internal shape is already encoded before pooling, through the sinusoidal position embeddings added to each patch feature (Sec.~\ref{sec:modeling}). The position given to the LLM therefore only needs to anchor the region in the image, which any position scheme can provide. We compare two such schemes at 1.7B (Table~\ref{tab:position-encoding}). Our default extends the LLM's native 1D RoPE to interleaved 3D M-RoPE and places each region token at the patch nearest the region's center. The alternative keeps 1D RoPE and adds an absolute position embedding computed from each region's mask, encoded via a learned linear projection. The two are within 0.3 points of each other on captioning, pointing, and counting, so \rvlm{} does not require a backbone with native M-RoPE support. We use M-RoPE because it adds no parameters, whereas the mask embedding does.}

\subsection{\texorpdfstring{\rtwo{Model Scale}}{Model Scale}}
\label{app:model-scale}
\rtwo{We perform the patch-versus-region comparison with a 1.7B LLM (Tables~\ref{tab:scale-stage1} and \ref{tab:scale-stage2}). After stage 1, \rvlm{}-1.7B improves pointing by 2.2 points and stays within 0.3 points on captioning and counting. After stage 2, it improves COCO and A-OKVQA by 0.7 to 1.0 points, and text-rich scores change by $-$1.1 to +1.0 points (DocVQA, ChartQA, InfoVQA).}

\subsection{\texorpdfstring{\rtwo{Text-Only and Video Evaluation}}{Text-Only and Video Evaluation}}
\label{app:text-video}
\rtwo{As shown in Table~\ref{tab:text-only}, region tokenization leaves text-only ability essentially unchanged (within 0.6 points).
As shown in Table~\ref{tab:video}, on video, which neither model is trained on, \rvlm{} stays within 0.5 points of \pvlm{} while using 3$\times$ fewer visual tokens. We leave a fuller development and exploration of region-based video-language models to future work.}

\section{Limitations}
\label{app:limitations}
\boldheader{Fine-grained visual content}
\rvlm{} underperforms \pvlm{} on InfoVQA ($-4.9$ points),
ScreenSpot Pro ($-3.3$), and ChartQA ($-0.8$). These tasks often require distinguishing very small text or interface elements, which may be absorbed into surrounding background regions during segmentation or patch assignment. Although complete patch coverage is retained, pooling can obscure these fine details. Our FH ablations (Appendix~\ref{app:fh-ablation}) show that text-rich tasks are sensitive to region granularity, suggesting that adjusting $\mathrm{min\_size}$, working resolution, or other FH parameters could mitigate these failures. We use a fixed configuration across tasks to develop a general-purpose VLM that balances performance and token efficiency. Adapting segmentation granularity to individual images or task requirements may further improve this trade-off but remains unexplored.

\ronetwo{\textbf{Text-coordinate grounding.}
\rvlm{}'s spatial precision relies on grounding-token decoding through the retained patch-to-region correspondence.
When coordinates are instead generated as text, region tokens trail patch tokens by 4.8 points on PixMo-Point (72.7 vs.\ 77.5 at 1.7B; App.~\ref{app:dense-outputs}), since a single region token can cover up to 16 patches.
VLMs that emit text coordinates would therefore need to adopt grounding-token decoding to benefit from region tokens. However, as shown by \citet{MolmoPoint}, this decoding is more accurate than text coordinates in its own right, while using 3 output tokens per point instead of 6--8 in text-based coordinate prediction.}

\rone{\textbf{Patch-based encoding.}
\rvlm{} changes the units the LLM consumes, but the vision encoder still processes the full patch grid, so its cost continues to scale with pixel area (App.~\ref{app:resolution-scaling}).
Encoders that operate natively on regions and skip redundant patches could extend the efficiency gains to the vision encoder.}

\boldheader{Architectural scope}
We develop and evaluate region tokenization within the MolmoPoint architecture~\citep{MolmoPoint}. Although the underlying approach is not specific to this model, we have not established its effectiveness across other VLM architectures. In particular, connectors that concatenate a fixed number of patch embeddings along the feature dimension cannot directly accommodate variable-sized regions without modification. Extending region tokenization to other connector designs and to long video remains future work.

\boldheader{Dependence on photometric boundaries}
FH constructs regions from local color differences, which may not align with semantic boundaries. This limitation may be particularly relevant for low-contrast or degraded imagery, including fog, camouflage, and motion blur, where meaningful structures have weak visual boundaries. We do not explicitly evaluate these conditions. Our limit of $w=16$ patches per region bounds the amount of content pooled into a single token, but it cannot recover boundaries missed during region construction.

\boldheader{Connector width trade-off}
We choose $w{=}16$ to balance visual-token reduction against connector padding. Although larger widths further shorten the LLM sequence, our structural ablation shows that they substantially increase the padded connector input (Appendix~\ref{app:connector-width-ablation}). The chosen width, therefore, represents a practical operating point rather than a universally optimal setting. More efficient handling of variable-sized regions may improve this trade-off.

\begin{figure}[t]
    \centering
    \includegraphics[width=\linewidth]{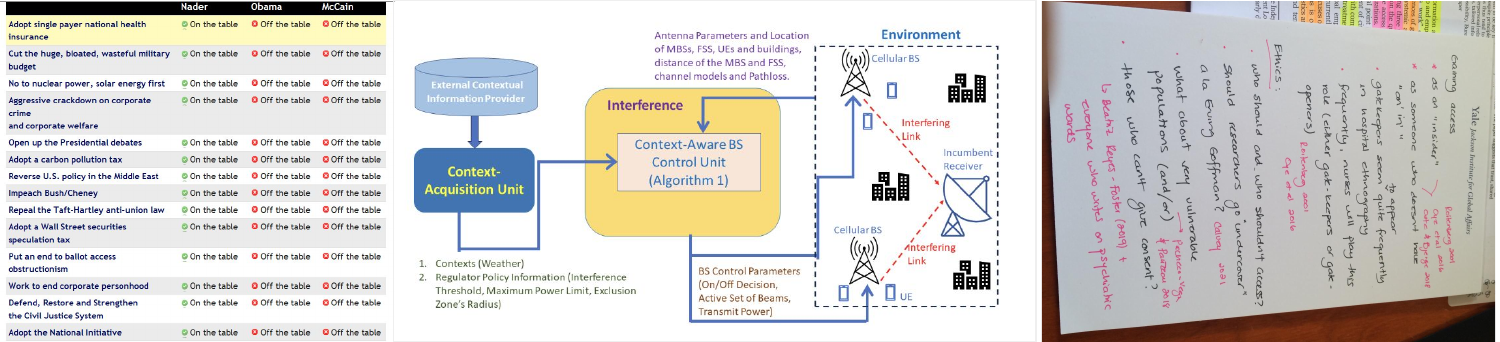}
    \vspace{-1.6em}
    \caption{{\bf Failure cases of T-REN-based \rvlm{}.} During VLM training, region tokens become spatially disconnected, making it difficult to associate text and visual elements with their correct locations. The model recognizes individual labels but misattributes them to neighboring rows, columns, or objects, producing incorrect descriptions of the image structure. For instance, in the first example, it incorrectly describes some initiatives as ``on the table'' for Obama, even though the image lists none. \pvlm{} and FH-based \rvlm{} do not show these failures.}
    \label{fig:disjoint-regions}
\end{figure}


%% file: tables/training-hparams.tex
\begin{table}[t]
    \caption{\textbf{Training hyperparameters for \rvlm{}.} Both stages jointly train the vision encoder, connector, and LLM. Stage 2 initializes from the final Stage-1 checkpoint.}
    \vspace{-0.6em}
    \label{tab:training-hparams}
    \centering
    \small
    \setlength{\tabcolsep}{12pt}
    \renewcommand{\arraystretch}{1.12}

    \begin{tabular}{@{}lcc@{}}
        \toprule
        \textbf{Setting}
        & \textbf{Stage 1}
        & \textbf{Stage 2} \\
        \midrule
        Training steps
            & $31{,}971$ & $24{,}000$ \\
        Global / microbatch size
            & $128$ / $8$ & $128$ / $8$ \\
        Packed sequence length
            & $3072$ & $3072$ \\
        Maximum training crops
            & $14$ & $14$ \\
        Maximum packed images
            & $28$ & $28$ \\
        Optimizer
            & AdamW & AdamW \\
        $\beta_1,\beta_2$
            & $(0.9,0.95)$ & $(0.9,0.95)$ \\
        $\varepsilon$
            & $10^{-6}$ & $10^{-6}$ \\
        Weight decay
            & $0$ & $0$ \\
        ViT learning rate
            & $1{\times}10^{-5}$ & $2{\times}10^{-5}$ \\
        LLM learning rate
            & $2{\times}10^{-5}$ & $2{\times}10^{-5}$ \\
        Connector learning rate
            & $2{\times}10^{-4}$ & $2{\times}10^{-5}$ \\
        ViT warmup
            & $2000$
            & $200$ \\
        LLM warmup
            & $2000$
            & $200$ \\
        Connector warmup
            & $200$
            & $200$ \\
        LR schedule
            & Cosine & Cosine \\
        Final LR fraction
            & $0.1$ & $0.1$ \\
        Gradient clipping
            & $1.0$ & $1.0$ \\
        \midrule
        Hardware
            & $16{\times}$ NVIDIA B300
            & $16{\times}$ NVIDIA B300 \\
        Distributed training
            & FSDP & FSDP \\
        Precision
            & bf16 & bf16 \\
        \bottomrule
    \end{tabular}
\end{table}

%% file: tables/training-mixtures.tex
\begin{table}[t]
    \caption{\textbf{Training data mixtures.} Patch and region models share these mixtures.}
    \vspace{-0.6em}
    \label{tab:training-mixtures}
    \centering
    \small
    \setlength{\tabcolsep}{5pt}
    \renewcommand{\arraystretch}{1.4}

    \begin{tabular}{ll}
        \toprule
        \textbf{Task family}
        & \textbf{Datasets} \\
        \midrule
        \rowcolor{black!8} \multicolumn{2}{c}{\bf Stage~1 (multimodal pre-training)} \\
        Captioning
            & PixMo-Cap \\
        OCR
            & olmOCR-mix \\
        Pointing / counting
            & PixMo-Points, PixMo-Count, CoSyn-point, COCO, LVIS \\
        Segmentation
            & COCO, LVIS \\
        Depth
            & \makecell[l]{SUN RGB-D, NYU Depth v2, DIODE, HM3D, DDAD, Waymo, Argoverse~2, \\ nuScenes, Aria Digital Twin, Aria Everyday Activities, Aria Nymeria, EgoExo4D, \\ Stereo4D} \\
        Camera
            & \makecell[l]{Stereo4D, SUN RGB-D, NYU Depth v2, EgoExo4D, DDAD, DIODE, \\ nuScenes, Argoverse~2, Waymo, ScanNet++, ARKitScenes, iBims-1, Aria Nymeria} \\
        Text-only NLP
            & Tulu-4 \\
        \midrule
        \rowcolor{black!8} \multicolumn{2}{c}{\bf Stage~2 (supervised fine-tuning)} \\
        Image QA
            & VQAv2, A-OKVQA (MC and DA), OK-VQA, ScienceQA, AI2D, TallyQA \\
        Captioning
            & PixMo-Cap, PixMo-CapQA, PixMo-AskModelAnything \\
        OCR
            & TextVQA, ChartQA, DocVQA, InfoVQA, ST-VQA, TabMWP \\
        Pointing / counting
            & PixMo-Points, PixMo-Count, CoSyn-point \\
        Segmentation
            & PixMo-Points, COCO, LVIS, EntitySeg \\
        Depth
            & \makecell[l]{SUN RGB-D, NYU Depth v2, DIODE, HM3D, DDAD, Waymo, Argoverse~2, \\ nuScenes, Aria Digital Twin, Aria Everyday Activities, Aria Nymeria, EgoExo4D, \\ Stereo4D} \\
        Text-only NLP
            & Tulu-4 \\
        \bottomrule
    \end{tabular}
\end{table}

%% file: tables/eval-tasks.tex
\begin{table*}[t]
    \centering
    \caption{\textbf{Our main evaluation suite}. Metrics match those reported in the main native-resolution comparison.
    \rtwo{The 35 tasks span natural images, documents, charts, diagrams, infographics, scene text, GUI screenshots, driving scenes, egocentric, exocentric, and stereo video frames, and indoor and outdoor RGB-D.}}
    \vspace{-0.6em}
    \label{tab:eval-tasks}
    \scriptsize
    \setlength{\tabcolsep}{6pt}
    \renewcommand{\arraystretch}{1.08}
    \begin{tabular}{@{}llp{8.1cm}@{}}
        \toprule
        \textbf{Dataset} & \textbf{Metric} & \textbf{Task} \\
        \midrule
        \rowcolor{black!8} \multicolumn{3}{c}{\textbf{Image QA}} \\
        A-OKVQA (DA)
          & VQA score
          & Open-ended knowledge-aware visual questions (direct answer). \\
        A-OKVQA (MC)
          & Accuracy
          & Multiple-choice knowledge-aware visual questions. \\
        AI2D
          & Accuracy (transparent)
          & Diagram science QA with multiple-choice answers. \\
        OK-VQA
          & VQA score
          & Open-ended visual questions requiring outside knowledge. \\
        ScienceQA
          & Accuracy
          & Multimodal science questions with images (image subset). \\
        VQAv2
          & VQA score
          & Open-ended natural-image visual question answering. \\
        ChartQA
          & Relaxed accuracy
          & Questions over charts and plots. \\
        DocVQA
          & ANLS
          & Questions over document images / scanned pages. \\
        InfoVQA
          & ANLS
          & Questions over infographics. \\
        TextVQA
          & VQA score
          & Questions that require reading text in natural images. \\
        \midrule
        \rowcolor{black!8} \multicolumn{3}{c}{\textbf{Pointing}} \\
        OSWorld-G
          & Overall accuracy
          & GUI grounding: point to UI elements from natural-language instructions. \\
        PixMo-Points
          & F1
          & Open-vocabulary pointing to all instances of a described object. \\
        PointBench
          & Average accuracy
          & Multi-skill pointing (spatial, counting, reasoning, steerable, etc.). \\
        SA-Co Gold
          & All-frequency F1
          & Long-tail concept pointing on SA-Co Gold (pointing split). \\
        ScreenSpot Pro
          & Overall accuracy
          & Professional desktop GUI grounding across application categories. \\
        ScreenSpot v2
          & Average accuracy
          & Mobile / desktop / web UI element grounding. \\
        \midrule
        \rowcolor{black!8} \multicolumn{3}{c}{\textbf{Counting}} \\
        CountBenchQA
          & Exact-count accuracy
          & Count objects described by a natural-language question. \\
        PixMo-Count
          & Exact-count accuracy
          & Count instances of a queried category in natural images. \\
        \midrule
        \rowcolor{black!8} \multicolumn{3}{c}{\textbf{Segmentation}} \\
        PixMo-Seg
          & CGF1
          & Open-vocabulary referring segmentation (mask prediction). \\
        RefCOCO
          & Overall IoU
          & Referring expression segmentation on COCO (RefCOCO). \\
        RefCOCO+
          & Overall IoU
          & Referring expression segmentation without location words. \\
        RefCOCOg
          & Overall IoU
          & Referring expression segmentation with longer expressions. \\
        SA-Co Gold
          & Mean CGF1 (group average)
          & Long-tail concept segmentation on SA-Co Gold (segmentation split). \\
        \midrule
        \rowcolor{black!8} \multicolumn{3}{c}{\textbf{Point-to-Depth}} \\
        Argoverse~2
          & Scale-invariant $\delta_{1.25}$
          & Estimate depth at queried 2D points in autonomous-driving scenes. \\
        Aria AEA
          & Scale-invariant $\delta_{1.25}$
          & Point-to-depth on Aria Everyday Activities egocentric video. \\
        Aria Nymeria
          & Scale-invariant $\delta_{1.25}$
          & Point-to-depth on Aria Nymeria egocentric recordings. \\
        DDAD
          & Scale-invariant $\delta_{1.25}$
          & Point-to-depth on Dense Depth for Autonomous Driving. \\
        DIODE
          & Scale-invariant $\delta_{1.25}$
          & Point-to-depth on Dense Indoor/Outdoor DEpth. \\
        EgoExo4D (ego)
          & Scale-invariant $\delta_{1.25}$
          & Point-to-depth from EgoExo4D egocentric cameras. \\
        EgoExo4D (exo)
          & Scale-invariant $\delta_{1.25}$
          & Point-to-depth from EgoExo4D exocentric cameras. \\
        nuScenes
          & Scale-invariant $\delta_{1.25}$
          & Point-to-depth in nuScenes driving scenes. \\
        NYU Depth v2
          & Scale-invariant $\delta_{1.25}$
          & Point-to-depth on indoor RGB-D scenes. \\
        Stereo4D
          & Scale-invariant $\delta_{1.25}$
          & Point-to-depth on internet stereo / video-derived depth. \\
        SUN RGB-D
          & Scale-invariant $\delta_{1.25}$
          & Point-to-depth on SUN RGB-D indoor scenes. \\
        Waymo
          & Scale-invariant $\delta_{1.25}$
          & Point-to-depth on Waymo Open Dataset driving scenes. \\
        \bottomrule
    \end{tabular}
\end{table*}

%% file: tables/fh-ablations.tex
\begin{table}[t]
    \centering
    \caption{{\bf FH parameter ablation.} Default settings are highlighted in blue. Token count is the mean visual-token count over the six tasks. NYUv2 reports scale-invariant $\delta_1$ using per-image median scale alignment.}
    \vspace{-0.6em}
    \label{tab:fh-ablation}
    \small
    \setlength{\tabcolsep}{2.4pt}
    \renewcommand{\arraystretch}{1.2}
    \begin{tabular}{lccccccc}
        \toprule
        \textbf{Setting}
          & \textbf{A-OKVQA}
          & \textbf{InfoVQA}
          & \textbf{PixMo-Point}
          & \textbf{PixMo-Count}
          & \textbf{RefCOCO}
          & \textbf{NYUv2}
          & \textbf{Token count} \\
        \midrule
        \rowcolor{black!8} \multicolumn{8}{c}{\textbf{\boldmath Scale $k$\quad($\mathrm{min\_size}=32$, $\mathrm{max\_side}=768$)}} \\
        $k{=}50$
          & 75.5 & 74.3 & 86.6 & 91.9 & 82.4 & 97.6 & 1171 \\
        \rowcolor{blue} $k{=}100$
          & 74.8 & 74.3 & 86.4 & 90.4 & 82.5 & 97.5 & 963 \\
        $k{=}200$
          & 75.4 & 72.8 & 85.8 & 90.9 & 82.2 & 97.5 & 794 \\
        $k{=}400$
          & 75.1 & 72.1 & 85.1 & 89.6 & 82.0 & 97.4 & 686 \\
        \midrule
        \rowcolor{black!8} \multicolumn{8}{c}{\textbf{\boldmath Minimum region size\quad($k=100$, $\mathrm{max\_side}=768$)}} \\
        $\mathrm{min\_size}{=}16$
          & 74.9 & 74.9 & 86.5 & 90.9 & 82.4 & 97.4 & 1032 \\
        \rowcolor{blue} $\mathrm{min\_size}{=}32$
          & 74.8 & 74.3 & 86.4 & 90.4 & 82.5 & 97.5 & 963 \\
        $\mathrm{min\_size}{=}64$
          & 75.5 & 72.2 & 85.8 & 90.6 & 82.4 & 97.4 & 860 \\
        $\mathrm{min\_size}{=}128$
          & 75.2 & 70.7 & 85.9 & 91.3 & 82.3 & 97.4 & 736 \\
        \midrule
        \rowcolor{black!8} \multicolumn{8}{c}{\textbf{\boldmath FH working resolution\quad($k=100$, $\mathrm{min\_size}=32$)}} \\
        $\mathrm{max\_side}{=}384$
          & 75.1 & 71.0 & 85.1 & 90.4 & 82.4 & 97.4 & 746 \\
        $\mathrm{max\_side}{=}512$
          & 75.0 & 71.6 & 85.9 & 91.1 & 82.5 & 97.4 & 830 \\
        \rowcolor{blue} $\mathrm{max\_side}{=}768$
          & 74.8 & 74.3 & 86.4 & 90.4 & 82.5 & 97.5 & 963 \\
        $\mathrm{max\_side}{=}1024$
          & 74.9 & 75.2 & 86.7 & 90.7 & 82.5 & 97.5 & 1060 \\
        \bottomrule
    \end{tabular}
\end{table}

%% file: tables/connector-width-ablation.tex
\begin{table}[t]
    \caption{\textbf{\boldmath Structural effect of connector width $w$.} Statistics are pooled over $1{,}480$ images from six tasks. Input memory estimates the storage of the padded bf16 connector input for one image; it excludes model parameters and intermediate activations.}
    \vspace{-0.6em}
    \label{tab:connector-width-ablation}
    \centering
    \small
    \setlength{\tabcolsep}{5.5pt}
    \begin{tabular}{lccccc}
        \toprule
        \textbf{\boldmath $w$} &
        \textbf{Visual tokens} &
        \textbf{Patches/token} &
        \textbf{Split \%} &
        \textbf{Padding} &
        \textbf{Input memory (MiB)} \\
        \midrule
        4 & 1,814 & 3.21 & 31.1\% & 18.4\% & 31.9 \\
        8 & 1,243 & 4.96 & 16.5\% & 40.4\% & 43.7 \\
        \rowcolor{blue} 16 & 980 & 6.63 & 7.8\% & 62.2\% & 68.9 \\
        32 & 863 & 7.79 & 3.5\% & 78.5\% & 121.3 \\
        64 & 812 & 8.43 & 1.5\% & 88.6\% & 228.4 \\
        128 & 791 & 8.73 & 0.7\% & 94.1\% & 444.8 \\
        \bottomrule
    \end{tabular}
\end{table}

%% file: tables/native-res-bucketted-results.tex
\begin{table}[t]
    \caption{{\bf PVLM vs.\ RVLM across five native-resolution ranges.} Task metrics are computed over all examples within each resolution range. A dash (--) indicates that the dataset contains no images in that range.}
    \vspace{-0.6em}
    \label{tab:native-res-bucketted-results}
    \centering
    \scriptsize
    \setlength{\tabcolsep}{2.5pt}
    \renewcommand{\arraystretch}{0.98}
    \begin{tabular}{l @{\cropgap} cc @{\cropgap} cc @{\cropgap} cc @{\cropgap} cc @{\cropgap} cc}
        \toprule
        \textbf{Task} 
        & \multicolumn{2}{c}{{\bf 0.14--0.41 MP}\;\;} & \multicolumn{2}{c}{{\bf 0.55--0.76 MP}\;\;} & \multicolumn{2}{c}{{\bf 0.83--1.38 MP}\;\;} & \multicolumn{2}{c}{{\bf 1.55--2.01 MP}\;\;} & \multicolumn{2}{c}{{\bf 2.08--2.63 MP}\;\;} \\
        \cmidrule(r){2-3}
        \cmidrule(r){4-5}
        \cmidrule(r){6-7}
        \cmidrule(r){8-9}
        \cmidrule(r){10-11}
        & \textbf{\pvlm{}} & \textbf{\rvlm{}} & \textbf{\pvlm{}} & \textbf{\rvlm{}} & \textbf{\pvlm{}} & \textbf{\rvlm{}} & \textbf{\pvlm{}} & \textbf{\rvlm{}} & \textbf{\pvlm{}} & \textbf{\rvlm{}} \\
        \midrule
        \rowcolor{black!8} \multicolumn{11}{c}{\textbf{All tasks}} \\
        \rowcolor{black!4} \textbf{Example-weighted average} & 85.3 & \perfsame{85.3} & 69.8 & \perfgain{70.2}{0.4} & 72.6 & \perfgain{73.8}{1.2} & 69.0 & \perfgain{73.6}{4.6} & 68.4 & \perfgain{73.7}{5.3} \\
        \rowcolor{black!4} \textbf{Family-balanced average} & 85.0 & \perfgain{85.3}{0.3} & 78.5 & \perfgain{79.4}{0.9} & 76.9 & \perfgain{77.9}{1.0} & 76.5 & \perfgain{81.9}{5.4} & 70.4 & \perfgain{79.4}{9.0} \\
        \rowcolor{black!4} \tval{\textbf{Token count}} & \tval{691} & \diff{660}{4.5} & \tval{987} & \diff{797}{19.3} & \tval{1,608} & \diff{1,076}{33.1} & \tval{2,499} & \diff{1,358}{45.6} & \tval{3,277} & \diff{1,581}{51.8} \\
        \midrule
        \rowcolor{black!8} \multicolumn{11}{c}{\textbf{Image QA}} \\
        A-OKVQA DA & 75.5 & 74.8 & -- & -- & -- & -- & -- & -- & -- & -- \\
        A-OKVQA MC & 88.6 & 88.2 & -- & -- & -- & -- & -- & -- & -- & -- \\
        AI2D & 94.9 & 94.5 & 96.7 & 94.8 & 95.0 & 98.3 & 96.2 & 97.4 & 96.0 & 96.0 \\
        ChartQA & 67.8 & 61.9 & 84.7 & 84.2 & 85.0 & 84.4 & 100.0 & 100.0 & 100.0 & 100.0 \\
        DocVQA & 97.2 & 99.3 & 95.8 & 92.6 & 89.8 & 92.3 & 94.4 & 90.4 & 93.2 & 92.9 \\
        OK-VQA & 65.8 & 66.0 & -- & -- & -- & -- & -- & -- & -- & -- \\
        ScienceQA & 98.4 & 98.0 & 97.6 & 97.3 & -- & -- & -- & -- & -- & -- \\
        TextVQA & 94.1 & 96.1 & 85.4 & 83.0 & 85.5 & 85.1 & -- & -- & 80.0 & 80.0 \\
        VQAv2 & 87.2 & 87.1 & -- & -- & -- & -- & -- & -- & -- & -- \\
        \rowcolor{black!4} \textbf{Task average} & 86.8 & 86.7 & 88.7 & 87.8 & 86.0 & 85.8 & 95.1 & 92.9 & 93.2 & 92.9 \\
        \rowcolor{black!4} \tval{\textbf{Token count}} & \tval{689} & \tval{662} & \tval{984} & \tval{612} & \tval{1,614} & \tval{1,098} & \tval{2,483} & \tval{1,260} & \tval{3,359} & \tval{1,636} \\
        \midrule
        \rowcolor{black!8} \multicolumn{11}{c}{\textbf{Pointing}} \\
        OSWorld-G & -- & -- & -- & -- & 71.7 & 73.8 & -- & -- & 43.6 & 43.3 \\
        PixMo Point & 81.2 & 81.6 & 90.6 & 89.5 & 86.0 & 86.6 & 83.1 & 87.3 & 88.1 & 94.0 \\
        PointBench & 71.0 & 71.6 & 56.7 & 56.9 & 68.1 & 65.2 & 62.5 & 64.9 & 59.8 & 58.7 \\
        SA-Co Gold & 91.2 & 91.6 & 88.6 & 89.3 & 90.4 & 91.5 & 87.6 & 90.6 & 78.3 & 87.4 \\
        ScreenSpot Pro & -- & -- & -- & -- & -- & -- & -- & -- & 29.3 & 26.0 \\
        ScreenSpot v2 & 33.3 & 33.3 & 30.4 & 30.7 & 31.7 & 31.7 & 30.6 & 28.0 & 79.8 & 82.8 \\
        \rowcolor{black!4} \textbf{Task average} & 90.4 & 90.8 & 86.9 & 87.5 & 88.9 & 89.8 & 83.7 & 86.7 & 68.5 & 74.1 \\
        \rowcolor{black!4} \tval{\textbf{Token count}} & \tval{719} & \tval{596} & \tval{990} & \tval{838} & \tval{1,620} & \tval{1,080} & \tval{2,497} & \tval{1,380} & \tval{3,279} & \tval{1,544} \\
        \midrule
        \rowcolor{black!8} \multicolumn{11}{c}{\textbf{Counting}} \\
        CountBenchQA & 90.8 & 91.8 & 87.2 & 87.2 & 94.9 & 94.9 & 100.0 & 100.0 & 61.5 & 92.3 \\
        PixMo Count & 91.9 & 91.9 & 84.4 & 93.3 & 89.9 & 91.3 & 74.5 & 90.9 & 61.9 & 87.3 \\
        \rowcolor{black!4} \textbf{Task average} & 90.9 & 91.8 & 85.7 & 90.5 & 91.7 & 92.6 & 75.3 & 91.1 & 61.9 & 87.8 \\
        \rowcolor{black!4} \tval{\textbf{Token count}} & \tval{552} & \tval{500} & \tval{1,004} & \tval{855} & \tval{1,649} & \tval{1,129} & \tval{2,729} & \tval{1,433} & \tval{3,354} & \tval{1,611} \\
        \midrule
        \rowcolor{black!8} \multicolumn{11}{c}{\textbf{Segmentation}} \\
        PixMo Seg & 53.9 & 52.5 & 67.8 & 65.0 & 57.1 & 58.1 & 58.9 & 65.9 & 48.6 & 70.8 \\
        RefCOCO & 82.0 & 82.5 & -- & -- & -- & -- & -- & -- & -- & -- \\
        RefCOCO+ & 78.5 & 78.7 & -- & -- & -- & -- & -- & -- & -- & -- \\
        RefCOCOg & 80.3 & 81.1 & -- & -- & -- & -- & -- & -- & -- & -- \\
        SA-Co Gold & 42.9 & 43.1 & 40.9 & 41.9 & 46.0 & 48.4 & 41.3 & 46.7 & 47.6 & 56.2 \\
        \rowcolor{black!4} \textbf{Task average} & 64.3 & 64.6 & 41.9 & 42.7 & 46.5 & 48.9 & 43.9 & 49.6 & 47.6 & 56.3 \\
        \rowcolor{black!4} \tval{\textbf{Token count}} & \tval{709} & \tval{669} & \tval{987} & \tval{843} & \tval{1,610} & \tval{1,077} & \tval{2,499} & \tval{1,386} & \tval{3,209} & \tval{1,596} \\
        \midrule
        \rowcolor{black!8} \multicolumn{11}{c}{\textbf{Depth}} \\
        Argoverse 2 & -- & -- & -- & -- & -- & -- & -- & -- & 77.9 & 81.8 \\
        Aria AEA & -- & -- & -- & -- & 68.5 & 69.5 & -- & -- & -- & -- \\
        Aria Nymeria & -- & -- & -- & -- & 54.6 & 55.6 & -- & -- & -- & -- \\
        DDAD & -- & -- & -- & -- & -- & -- & -- & -- & 81.7 & 87.7 \\
        DIODE & -- & -- & -- & -- & 92.2 & 94.0 & -- & -- & -- & -- \\
        EgoExo4D Ego & 89.1 & 88.9 & -- & -- & -- & -- & -- & -- & -- & -- \\
        EgoExo4D Exo & -- & -- & 84.0 & 82.9 & -- & -- & -- & -- & -- & -- \\
        nuScenes & -- & -- & -- & -- & -- & -- & 84.7 & 88.8 & -- & -- \\
        NYU Depth v2 & 97.8 & 97.5 & -- & -- & -- & -- & -- & -- & -- & -- \\
        Stereo4D & 89.0 & 88.6 & -- & -- & -- & -- & -- & -- & -- & -- \\
        SUN RGB-D & 97.5 & 97.7 & 99.0 & 98.8 & -- & -- & -- & -- & -- & -- \\
        Waymo & -- & -- & -- & -- & -- & -- & 84.2 & 89.8 & 83.9 & 91.9 \\
        \rowcolor{black!4} \textbf{Task average} & 92.8 & 92.6 & 89.2 & 88.5 & 71.4 & 72.7 & 84.5 & 89.1 & 80.6 & 86.1 \\
        \rowcolor{black!4} \tval{\textbf{Token count}} & \tval{634} & \tval{542} & \tval{955} & \tval{734} & \tval{1,372} & \tval{821} & \tval{2,321} & \tval{1,125} & \tval{3,254} & \tval{1,533} \\
        \bottomrule
    \end{tabular}
\end{table}

%% file: tables/inference-cost-sweep.tex
\begin{table}[t]
    \caption{{\bf Inference cost across image resolutions.} Generation latency, end-to-end (E2E) latency, and peak GPU memory for \pvlm{} and \rvlm{} at batch sizes 1 and 8. Measurements use 512 randomly sampled PixMo-Cap examples on a single NVIDIA H100, with 64 warm-up batches and five runs. Generation is capped at one output token. Latencies are reported as mean $\pm$ standard deviation. While \pvlm{} is slightly faster at low crop counts, \rvlm{} achieves increasing latency and memory savings as image resolution grows.}
    \vspace{-0.6em}
    \label{tab:inference-cost-sweep}
    \centering
    \small
    \setlength{\tabcolsep}{7pt}
    \renewcommand{\arraystretch}{1.15}
    \begin{tabular}{c cc cc cc}
        \toprule
        \textbf{Crops}
        & \multicolumn{2}{c}{\textbf{Generation latency\ (ms)}}
        & \multicolumn{2}{c}{\textbf{E2E latency (ms)}}
        & \multicolumn{2}{c}{\textbf{Peak memory (GiB)}} \\
        \cmidrule(lr){2-3}\cmidrule(lr){4-5}\cmidrule(lr){6-7}
        & \textbf{\pvlm{}} & \textbf{\rvlm{}}
        & \textbf{\pvlm{}} & \textbf{\rvlm{}}
        & \textbf{\pvlm{}} & \textbf{\rvlm{}} \\
        \midrule
        \rowcolor{black!8} \multicolumn{7}{c}{\textbf{Batch size 1}} \\
        1 & $37.9{\pm}1.5$ & $40.3{\pm}1.9$ & $38.6{\pm}1.5$ & $41.1{\pm}1.9$ & 19.23 & 19.43 \\
        4 & $45.4{\pm}1.1$ & $48.5{\pm}1.6$ & $46.2{\pm}1.1$ & $49.4{\pm}1.6$ & 19.45 & 19.70 \\
        8 & $61.7{\pm}0.8$ & $55.2{\pm}1.2$ & $63.0{\pm}0.8$ & $56.1{\pm}1.2$ & 19.76 & 19.64 \\
        12 & $76.7{\pm}0.8$ & $63.9{\pm}1.6$ & $77.8{\pm}0.8$ & $64.9{\pm}1.6$ & 20.09 & 19.79 \\
        16 & $92.0{\pm}0.9$ & $73.8{\pm}1.4$ & $93.0{\pm}0.9$ & $74.9{\pm}1.4$ & 20.36 & 19.96 \\
        24 & $127.8{\pm}1.2$ & $88.9{\pm}1.3$ & $129.0{\pm}1.2$ & $90.1{\pm}1.3$ & 20.98 & 19.95 \\
        32 & $162.1{\pm}1.4$ & $105.0{\pm}1.3$ & $163.3{\pm}1.4$ & $106.2{\pm}1.3$ & 21.60 & 20.02 \\
        \midrule
        \rowcolor{black!8} \multicolumn{7}{c}{\textbf{Batch size 8}} \\
        1 & $13.3{\pm}0.0$ & $14.5{\pm}0.1$ & $13.4{\pm}0.0$ & $14.7{\pm}0.1$ & 20.86 & 22.00 \\
        4 & $24.5{\pm}0.1$ & $24.4{\pm}0.1$ & $24.7{\pm}0.1$ & $24.6{\pm}0.1$ & 22.63 & 24.20 \\
        8 & $39.2{\pm}0.1$ & $30.8{\pm}0.1$ & $39.4{\pm}0.1$ & $31.0{\pm}0.1$ & 25.03 & 23.72 \\
        12 & $55.1{\pm}0.1$ & $40.9{\pm}0.1$ & $55.3{\pm}0.1$ & $41.2{\pm}0.1$ & 27.44 & 24.98 \\
        16 & $71.2{\pm}0.2$ & $50.9{\pm}0.1$ & $71.5{\pm}0.2$ & $51.2{\pm}0.2$ & 29.86 & 26.30 \\
        24 & $106.4{\pm}0.5$ & $65.8{\pm}0.2$ & $106.8{\pm}0.5$ & $66.2{\pm}0.3$ & 34.70 & 26.94 \\
        32 & $144.6{\pm}0.4$ & $83.7{\pm}0.4$ & $145.0{\pm}0.3$ & $84.2{\pm}0.4$ & 39.57 & 30.15 \\
        \bottomrule
    \end{tabular}
\end{table}